\documentclass[10pt,journal,compsoc]{IEEEtran}
\usepackage{graphicx}
\usepackage{stfloats}
\usepackage[nocompress]{cite}
\usepackage{amsmath}
\usepackage{cases}
\usepackage{algorithmic}
\usepackage[caption=false,font=footnotesize,labelfont=sf,textfont=sf]{subfig}
\usepackage{xcolor}
\usepackage{amsfonts}
\usepackage{soul}
\usepackage{hyperref}
\ifCLASSINFOpdf
\else
\fi
\begin{document}
%
\title{A Survey on Class Imbalance in Federated Learning}
%
%
%
%

\author{Jing Zhang,
        Chuanwen~Li,~\IEEEmembership{Member,~IEEE,}
        Jianzgong~Qi,~\IEEEmembership{?}
        and~Jiayuan~He~\IEEEmembership{?}
\IEEEcompsocitemizethanks{\IEEEcompsocthanksitem J. Zhang and C. W. Li were with the School of Computer Science and Engineering, Northeastern University, Shenyang 110169, China.\protect\\
E-mail: lichuanwen@mail.neu.edu.cn
\IEEEcompsocthanksitem J. Qi is with School of Computing and Information Systems, The University of Melbourne, Australia.
\IEEEcompsocthanksitem J. He are with School of Computing Technologies, RMIT University, Australia.}
\thanks{Manuscript received April 19, 2005; revised August 26, 2015.}}

%
%

\markboth{Journal of \LaTeX\ Class Files,~Vol.~14, No.~8, August~2015}%
{Shell \MakeLowercase{\textit{et al.}}: Bare Demo of IEEEtran.cls for Computer Society Journals}
%



\IEEEtitleabstractindextext{%
\begin{abstract}		
	Federated learning, which allows multiple client devices in a network to jointly train a machine learning model without direct exposure of clients' data, is an emerging distributed learning technique due to its nature of privacy preservation. However, it has been found that models trained with federated learning usually have worse performance than their counterparts trained in the standard centralized learning mode, especially when the training data is imbalanced. In the context of federated learning, data imbalance may occur either locally one one client device, or globally across many devices. The complexity of different types of data imbalance has posed challenges to the development of federated learning technique, especially considering the need of relieving data imbalance issue and preserving data privacy at the same time. Therefore, in the literature, many attempts have been made to handle class imbalance in federated learning. In this paper, we present a detailed review of recent advancements along this line. We first introduce various types of class imbalance in federated learning, after which we review existing methods for estimating the extent of class imbalance without the need of knowing the actual data to preserve data privacy. After that, we discuss existing methods for handling class imbalance in FL, where the advantages and disadvantages of the these approaches are discussed. We also summarize common evaluation metrics for class imbalanced tasks, and point out potential future directions. 

\end{abstract}

\begin{IEEEkeywords}
Federated learning, class imbalance, privacy protection, classification.
\end{IEEEkeywords}}

\maketitle

\IEEEdisplaynontitleabstractindextext

%
\IEEEpeerreviewmaketitle

\IEEEraisesectionheading{\section{Introduction}\label{sec:introduction}}

\IEEEPARstart{T}he performance of machine learning (ML) models is highly reliant on the volume and quality of the data that they are trained on. The classical training strategy of ML models requires the training data to be hosted at one place, meaning data collected at different devices needs to be transferred to a centralized server, which poses privacy, security, and processing risks. Federated learning (FL) has been identified as a promising technique to address these risks~\cite{mcmahan2017communication}\cite{yin2021comprehensive}. FL is a distributed training framework, which allow edge devices (e.g., laptops and mobiles at client user ends) to collaboratively train a ML model without exposing the local data on one edge device to other devices or the central server. 

In general, FL employs a set of heterogeneous edge devices and a central server to coordinate the learning process. To avoid data transfer across devices, the data collected by one edge device is preserved locally and is used to train a local model. Instead of transferring data directly, the edge devices transfer trained models to the central server, which are then aggregated into a global model. The decentralized architecture of FL poses several challenges in model training, including communication cost, system heterogeneity, and statistical heterogeneity \cite{li2020federated}. In particular, the statistical heterogeneity refers to that training data is not independently and identically (non-IID) distributed across edge devices in the network \cite{li2021survey} \cite{truex2019hybrid}. Since optimization algorithms usually assume training data to be IID, studies have shown that statistical heterogeneity may lead to slow convergence, sometimes even divergence, and non-trivial performance deterioration in model training.

\begin{figure}[htbp]
	\centering
	\subfloat[Classification accuracy.]{\includegraphics[scale=0.36]{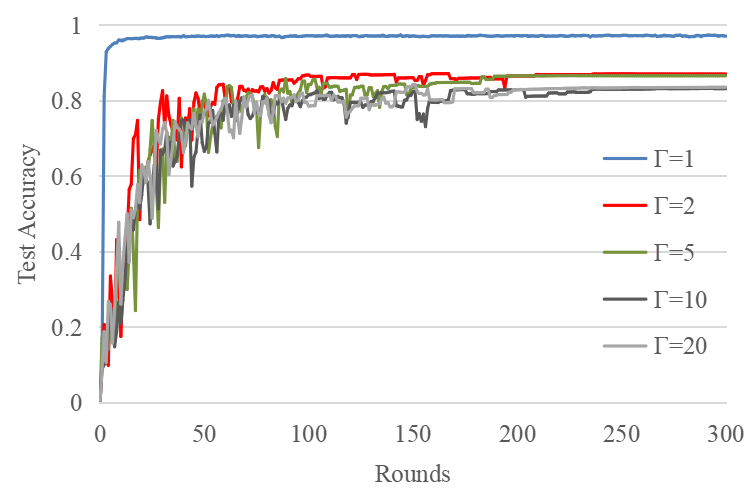}
		\label{fig_second_case}}
	\hfil
	\subfloat[Class distribution.]{\includegraphics[scale=0.4]{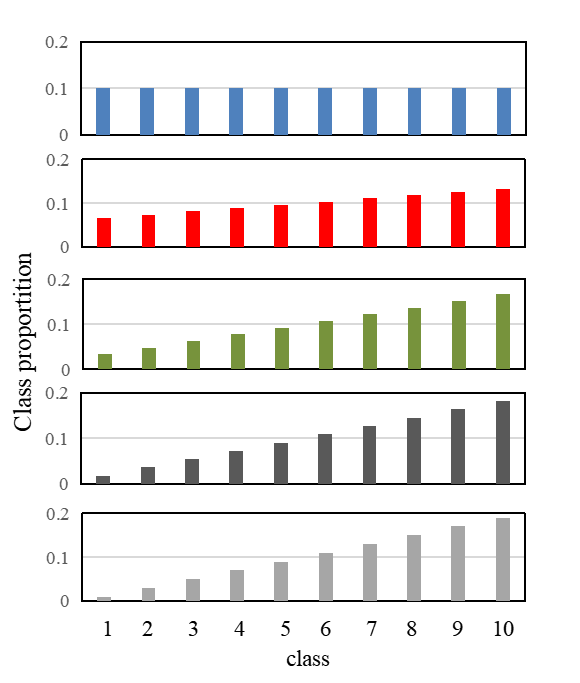}
		\label{fig_first_case}}	
	\caption{Example of class imbalanced data.}
	\label{Example of imbalanced data}
\end{figure}

In this paper, we focus on a special type of statistical heterogeneity \cite{li2021fedrs}, i.e., data with class imbalance, in FL. Class imbalance is also referred as label skew, class unbalance, and data imbalance in the literature \cite{tarekegn2021review} \cite{branco2016survey} \cite{kaur2019systematic} \cite{haixiang2017learning} \cite{krawczyk2016learning}. Fig. \ref{Example of imbalanced data} illustrates an example of the adversarial impact of class imbalance on model training. We use MNIST dataset (with 10 classes) to generate datasets with different degrees of class imbalance and then use these datasets for model training. We use $\Gamma$ to denote the degree of class imbalance, i.e., the size ratio of the largest class and the smallest class, where $\Gamma=1$ indicates balanced class distribution and higher value of $\Gamma$ indicates more severe class imbalance. 

We see that the classification accuracy is approximately 97\% when $\Gamma=1$. When the class imbalance degree is between 2 and 5, classification accuracy drops 86\%, and when the class imbalances degree is between 10 and 20, the classification accuracy further drops to 83\%. Such biased sample spaces are found in many real-world applications of FL. For example, in medical domain, the number of patients diagnosed with different diseases varies greatly \cite{dong2019semantic} with only very few cases for rare diseases.

As in conventional ML, class imbalance may cause significant performance degradation of FL models, especially for the minority classes \cite{mou2021optimized} \cite{li2021sample} \cite{duan2020self} \cite{zhang2021dubhe}. However, in many practical situations, the minority classes cannot be overlooked and sometimes even play a more important role than the majority classes. Consider a FL framework for health tracking, where multiple wearable devices are deployed for joint model training. These participating wearable devices should be more sensitive to abnormal heart rates than normal scenarios, although normal heart rates are dominant in the observation data. 

There are five main reasons for classification performance degradation of ML methods on class imbalanced datasets \cite{haixiang2017learning}: (1) standard classifiers are designed assuming balanced training data; (2) the classification process is guided by global performance metrics like classification accuracy which induces a bias towards samples of majority class; (3) rare minority class samples may potentially be treated as noises by the classifier; (4) minority samples may overlap with samples from other classes where the prior probabilities of these classes are equal; and (5) small disjuncts, lack of density, small sample sizes, and high feature dimensionality make the learning of minority class more challenging.

Handling class imbalance has been a long-standing research problem in ML with numerous approaches being proposed and evaluated \cite{krawczyk2016learning}. Most approaches address the class imbalance via resolving the aforementioned root causes of performance degradation, e.g., designing classifiers that are more sensitive to minority classes and resampling training data to generate balanced sample space \cite{galar2011review} \cite{krawczyk2014cost} \cite{loyola2016study}. 

In the context FL, handling class imbalance is even more challenging mainly for two reasons. First, due to the de-centralized nature of FL, class imbalance may occur at different levels, including local level on one or a few client devices and global level which is a systematic imbalance issue across many clients. Moreover, the mismatch in class distributions between the clients and the server may also cause performance degradation in FL. It has been demonstrated mathematically that the class imbalance in training data can reduce the classification performance of FL applications \cite{duan2019astraea}. In particular, experimental results show that class imbalance that occurs on the global level will be detrimental to the performance globally \cite{xiao2021experimental} \cite{chou2022grp}, and the differences in class distributions amongst all clients participating in FL will lower the overall performance and delay convergence of the global model \cite{sittijuk2021performance} \cite{diwangkara2020study}. 

Second, preserving user privacy is among the prime priorities in FL systems. Therefore, disclosing clients' data, even just the class distributions of clients' data, may lead to potential risks in leaking users' confidential information. As a result, managing class imbalance becomes much more difficult since judgments need to be made without having access to specific information about how severe the class imbalance is in the training data. Therefore, many approaches for handling class imbalance in FL systems follow a two-stage process: (1) class distribution estimation; and (2) addressing class imbalance in FL. 

Due to the rapid development of FL, there exist several survey papers focusing on different aspects of FL. In \cite{zhang2021survey} \cite{yang2019federated}, a general overview of FL and its applications is provided. In-depth explanations of advances and challenges of FL can be found in \cite{kairouz2021advances} \cite{khan2021federated}. Threat analyses and more privacy-preserving methods in FL are presented in \cite{yin2021comprehensive} \cite{mothukuri2021survey}. 

The class imbalance problem has also attracted lots of attention as it is prevalent in many real-life applications. There are also several survey papers on the class imbalance issue in traditional ML. In \cite{kaur2019systematic} \cite{haixiang2017learning} , the problem of class imbalance emerging around all real-world applications is inspected and an overview for the state-of-the-art solutions for the class imbalance problem in traditional ML is provided. There also exist some survey papers that provide reviews of approaches for handling non-IID data in FL. Non-IID data refers to the statistical heterogeneity found in training data, which may cause model degradation in FL \cite{zhao2018federated}. In \cite{zhu2021federated}, the impact of non-IID data on FL is examined in great detail, and a review of recent research on non-IID data problem in FL model is provided. 

Although class imbalance is a special type of non-IID data, the general approaches for non-IID data usually cannot be directly applied to address the problem class imbalance in FL. This is because most methods for non-IID data focus on achieving better and faster convergence of the global model in FL. This differs approaches for addressing class imbalance, which mainly aim to improve the classification performance of minority classes. None of the aforementioned survey papers have provided systematic study on the impact of class imbalance in FL, and none of these existing works have reviewed approaches that focusing on addressing the class imbalance problem in FL.

In this paper, we present a comprehensive review of the class imbalance problem in FL. Starting from a summary of FL structures and different types of class imbalance in FL systems, we review and discuss recent advancements in addressing class imbalance in FL, where we first review the existing methods for estimating class distributions and then systematicly discuss existing approaches that deal with class imbalanced learning in FL. We also summarize the metrics that are used in the literature for evaluating the classification performance of FL models, considering the class imbalance issue. To the best of our knowledge, this is the first paper that provides a systematic and focused review over the problem of class imbalance in FL. 


The rest of this paper is organized as follows. Section~\ref{sec:preliminaries} introduces the definition FL in general, and discusses various types of class imbalance in FL. Section~\ref{sec:class_estimation} reviews existing methods for estimating class distribution in FL, and Section~\ref{sec:classification} reviews approaches for addressing class imbalance problem in FL. Section~\ref{sec:metric} summarizes the performance metrics for the classification models in FL. Section~\ref{sec:challenges} discusses challenges and future directions. Finally, we conclude this paper with Section~\ref{sec:conclusion}.
\vspace{-0.2cm}

\section{Preliminaries}
\label{sec:preliminaries}
We provide an introduction to FL in this section by first explaining the concept and then classifying FL techniques based on data partitioning.
\vspace{-0.2cm}
\subsection{Background}
The remarkable success of ML in a wide range of applications (e.g., healthcare, Internet-of-things, and e-commerce) has proven the effectiveness of data-driven approaches. As these approaches are data-driven, access to high-quality and large volume of training data is one of the key factors to enhance model performances. However, accessing user-generated contents (e.g., patients' health records, sensors' readings, and users' browse history), which raises concerns in data security and user privacy. Therefore, federated learning (FL), first proposed by Google, has quickly become a popular framework for practical ML applications, where user privacy is a primary concern. FL is a de-centralized strategy for training ML models. It allows edge devices to collaboratively train a ML model, while keeping the data privacy of each edge device. As such, FL has the potential of bringing more user devices into the training process, leading to a more effective global model via combining the data and computational resources of user/edge devices.

A closely related concept to FL is distributed ML \cite{kim2016deepspark} \cite{sergeev2018horovod}. In distributed ML, the whole training dataset is divided into many smaller partitions, where each partition is transferred to one computing node. The aim of distributed ML is to share the training cost among computing nodes, and hence, data sharing across nodes is permitted if necessary. This is difference from FL, where preserving data privacy is the top priority task.

\begin{figure*}[!t]
	\centering
	\setlength{\abovecaptionskip}{-0.3cm}
	\includegraphics[scale=0.65]{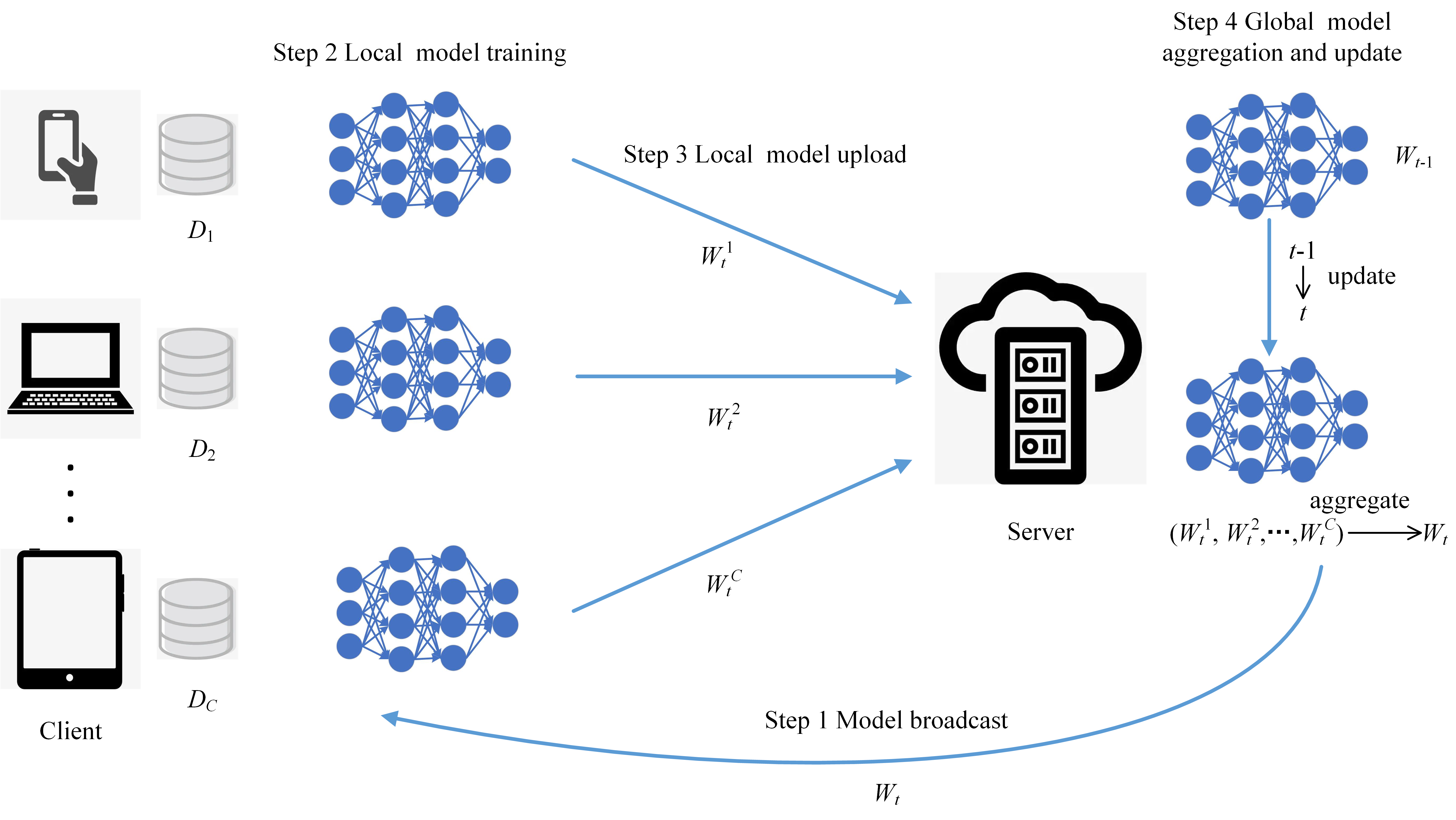}
	\caption{A general framework of federated learning. }
	\label{Framework of FL}
\end{figure*}

We depict a general framework of federated learning in Fig. \ref{Framework of FL}. FL assumes $C$ independent client devices (e.g., mobile phones, tablets, and laptops of end users) that will participate in the model training and one central server for coordinating the training. Let $c\in [C] = \{1, 2,\ldots,C\}$ be one participating device, and $ D_{c} = \left\lbrace  \left( x_{1}, y_{1} \right) , \dots  , \left( x_{N_c}, y_{N_c} \right)  \right\rbrace  $ be the private dataset owned by client $c$, where $N_c$ represents the number of training instances of client $c$, and $ (x_{i}, y_{i}) $ represents the $i$-th training instance with input of $x_{i}$ and target class label of $y_i$. The overall training process consists of four steps. The central server maintains a global model. At the beginning, this global model is copied to each participating client. Then each client will use their own data to update the global model. The trained local models are then uploaded to a central server, where all trained models are aggregated into a new global model. It is usually assumed that all participants including the server are semi-honest, i.e., they follow the exact protocols of FL but may be curious about other data \cite{li2021sample}. 

The FedAvg \cite{mcmahan2017communication} is the first and most famous FL model. It is an iterative algorithm of many rounds that gradually optimizes an objective function until convergence. Specifically, the objective of $t$-th training round is formulated as:
\begin{equation}
	\min _{W} f(W)=\sum_{i=1}^{K} \frac{N_i}{N} F_{i}(W_{t}^{i}) 
\end{equation}
where $ K $ is the number of clients that participate in the $t$-th round, $N=\sum_{i=1}^K N_i$ represents the total number of instances in the $t$-th round, and $ F_{i}(W) $ represents the loss function on the $i$-th participating client in the $t$-th round.

One training round in FedAvg can be summarized as follows:
\begin{itemize}
	\item[1] The server randomly selects a subset of clients and distributes the global model to these selected clients.
	\item[2] Each selected client in this round updates the received model by continual training it using their local dataset.
	\item[3] Each client that participate in this round sends their updated model back to the server.
	\item[4] The server aggregates the received clients' models into a new global model.
\end{itemize}

At $t$-th round, the central server aggregates local models uploaded by clients as follows:
\begin{equation}
	W_{t}= \sum_{i=1}^{K} \frac{N_{i}}{N} W_{t}^{i} 
\end{equation}
where $W_{t}^{i}=W_{t-1}-\eta g_{i}$, $ g_{i}=\bigtriangledown  F_{i}(W_{t}^{i}) $ and $  \eta $  is a fixed learning rate.

In addition to iterative aggregation, FedAvg optimizes communication efficiency by executing more local updates and fewer global updates. With these, FedAvg was demonstrated experimentally to function successfully with non-IID data \cite{nguyen2020efficient}. It is shown that the CNN model trained by FedAvg can achieve 99 percent test accuracy on non-IID MNIST dataset.

Inspired by the success of FedAvg, many more techniques for FL have been proposed \cite{zhang2021survey}, with the effectiveness and accuracy rapidly approaching their counterpart models that are trained in a traditional centralized manner \cite{konevcny2016federated}. FL is playing a more and more important role in many privacy-preserved ML applications.
\vspace{-0.2cm}

\subsection{Categorization Of Federated Learning Models}
Given a dataset, we define its feature space as the set of attributes that is used to describe data samples. We further define its sample space as the source entities from which samples are collected. For example, for a dataset containing patients' health records, the set of clinical tests that these patients have completed is the feature space of the dataset, and the set of patients represents the sample space.

In FL, datasets are collected and owned by different client devices. Therefore, the characteristics of these datasets may vary significantly, resulting in different distribution patterns of feature and/or sample space. Following Yang et al. \cite{yang2019federated}, we divide FL into three categories based on the types of mismatches amongst client datasets: (1) horizontal FL; (2) vertical FL; and (3) federated transfer learning~\cite{zhang2021survey} \cite{yang2019federated} \cite{khan2021federated}. 

\begin{figure*}[!t]
	\centering 	
	\subfloat[Horizontal FL.]{\includegraphics[scale=0.35]{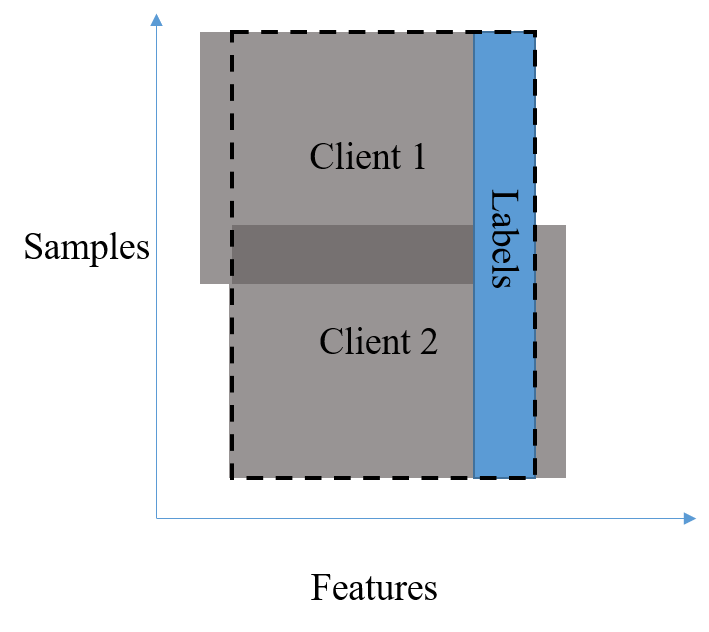}
		\label{fl_horizontal_case}}
	\hfil
	\subfloat[Vertical FL.]{\includegraphics[scale=0.35]{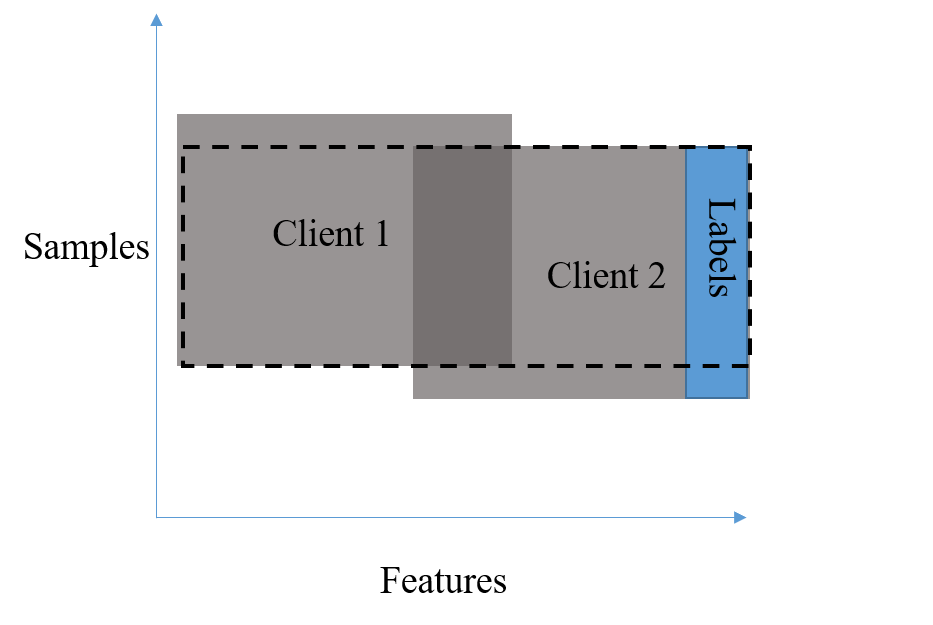}
		\label{fl_vertical_case}}
	\hfil
	\subfloat[Federated transfer learning.]{\includegraphics[scale=0.35]{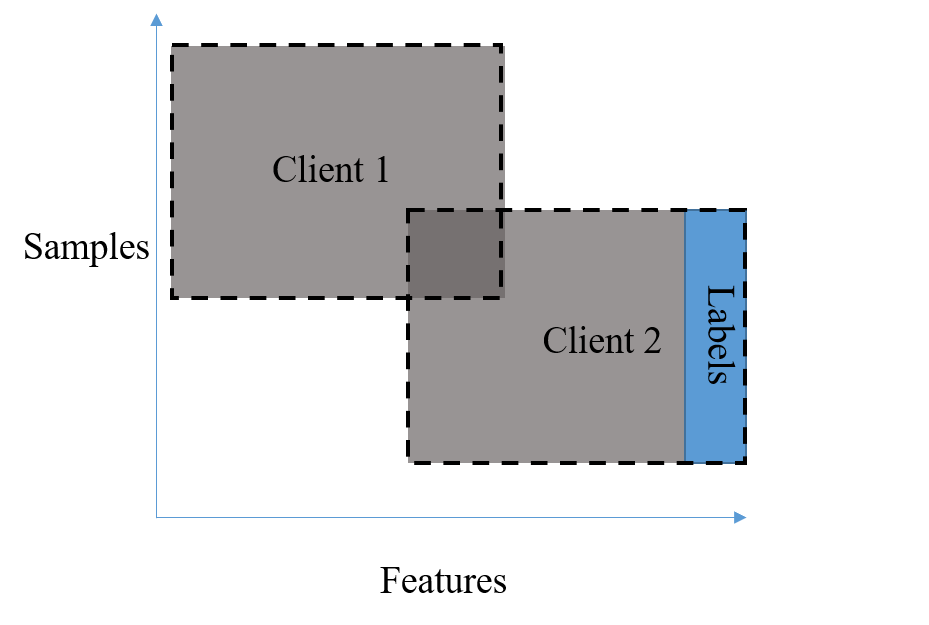}
		\label{fl_transfer_case}}
	\caption{Categorization of FL models based on feature and sample distributions of client datasets.}
	\label{Type}
\end{figure*}

\textbf{Horizontal FL}. In horizontal FL, client datasets across edge devices are highly similar in feature space but are different in sample space. That is, the client datasets on different devices can be divided by the sample dimension: one client dataset has a similar set of features as that of another client, and is an extension of the other dataset in terms of samples. To give an example, suppose we have multiple pathology sites that provide similar clinical test services in different regions. Hence, the data collected by these pathology sites will likely have different groups of patients (i.e., sample space) but share the same clinical tests (i.e., feature space). In this regard, horizontal FL benefits a ML model in terms of expanding the sample size of training data. In the literature, model aggregation in horizontal FL mainly focuses on gradient aggregation, i.e., all clients compute and upload local gradients based on their local datasets to the server for aggregation. It has been pointed out that private client information may still be leaked during communication of gradients in the horizontal FL. Homomorphic encryption \cite{yuan2013privacy}, differential privacy \cite{huang2020dp}, and safe aggregation \cite{mohassel2017secureml} have been proposed to address this issue.

\textbf{Vertical FL.} In vertical FL, client datasets have strong overlap in sample space but are different in feature space. This means sample entities are similar across client datasets, and each client dataset extends the features of these samples. An example of FL could be multiple devices storing the clinical test results of the same patients in different years. Here, the sample entities (i.e., patients) are identical on different devices but the feature spaces are different (i.e., different years of health records). To train a global model in vertical FL, it is critical to encrypt different features on different devices. Feature encryption has been successfully been applied in developing federated systems based on various ML models, including logical regression \cite{he2021secure}, tree structure model \cite{chen2021fed}, and neural networks \cite{dai2021vertical}.

\textbf{Federated transfer learning.} In federated transfer leaning, client datasets do not have much overlaps in either sample or feature space. In the health records example, suppose the healthcare providers offer different services and are located in different regions. Hence, the datasets owned by these healthcare providers will be different in feature space (i.e., different healthcare services) and sample space (i.e., different patients). In this case, there are no obvious overlaps between client datasets. However, the inherent knowledge carried by one dataset may still be beneficial for us to understand another client dataset. Thus, many efforts have been made to develop transfer learning techniques to extract generalizable features that can be transferred to other datasets~\cite{zhang2022transfer}. The confidentiality of client data can be preserved through the federated transfer learning. Moreover, it can transfer the model of auxiliary tasks to director learning, thus solving the problem of the small amount of data.

In Fig.~\ref{Type}, we illustrate the three types of FL, where a grey rectangle represents the feature and sample space of a client dataset. The dashed rectangle in the figure represents data for FL. We highlight the labels provided by client datasets in blue. In vertical FL, additional client datasets are only used to enhance the features of data samples, therefore only one client dataset is required to provide target class labels. The same applies to federated transfer learning where additional client datasets are usually on different tasks and are used to provide external knowledge. Thus, the class imbalance issue in vertical FL and federated transfer learning only occurs if the client dataset that provides labels has class imbalance. This is similar to class imbalance in traditional non-federated models, which has been covered by many literature. Therefore, in this paper, we mainly focus on class imbalance issue in horizontal FL.

\subsection{Class Imbalance In Federated Learning}
The problem of imbalanced data is a common issue in many real-world datasets \cite{choudhury2019predicting} \cite{xiao2021experimental}. FL is especially prone to class imbalance due to its de-centralized structure. It has been demonstrated that classification performance will degrade significantly under severe class imbalance in federated systems developed for various applications including healthcare \cite{li2021fedrs} \cite{hahn2019privacy} \cite{huang2019patient} \cite{hauschild2022federated}, anomaly detection \cite{shingi2020federated} \cite{cheng2022class} \cite{cheng2022blockchain} \cite{geng2022bearing}, fraud detection \cite{zheng2021federated} \cite{yang2019ffd}, mobile application\cite{duan2019astraea} \cite{duan2020self}, Internet-of-Things \cite{nguyen2020efficient} \cite{yu2020learning},  unmanned aerial vehicles \cite{mrad2021federated}, audio data \cite{green2021federated}, and intelligent control \cite{hua2020blockchain}. 

Since FL has a hierarchical structure of training with one centralized server governing a set of client devices, class imbalance in FL may occur at lower level on client datasets and also at global level. Next, we introduce the categorization of class imbalance in FL that is commonly used in the literature \cite{duan2019astraea}. 

For ease of presentation, we denote the number of data samples in class $ p $ on client $ i $ by $ N _{i}^{p} $. 
\begin{itemize}
	\item[1)] \textbf{Local imbalance}: the local dataset owned by each client is imbalanced. The local imbalance $ \gamma_{i} $ for client $ i $ is the ratio between the sample number of the majority class on client $ i $ and the sample number of the minority class on client $ i $, i.e., $ \gamma_{i} = max_{p}\left\lbrace N _{i}^{p}\right\rbrace / min_{p}\left\lbrace N _{i}^{p}\right\rbrace  $. In extreme situations, it is possible that $ min_{p}\left\lbrace N _{i}^{p}\right\rbrace=0 $ on one client dataset. 
	
	\item[2)] \textbf{Global imbalance}: the union of clients' datasets (i.e., global dataset) is imbalanced. From the global perspective, the global class imbalance $ \Gamma$ is defined as the ratio between the total majority class sample number across all clients and that of the minority class, $ \Gamma = max_{p}  \left\lbrace {\textstyle \sum_{i}^{}}N _{i}^{p}\right\rbrace / min_{p}\left\lbrace {\textstyle \sum_{i}^{}}N _{i}^{p}\right\rbrace  $.
	
	\item[3)] \textbf{Mismatch imbalance}: each client's local imbalance differs from the global imbalance. In reality, such a mismatch is found to be common and significant. One class may be the majority class with client $i$ but become the minority class on the global level. To quantify this mismatch more accurately, we use vector $ v_{i} = [N _{i}^{1},..., N _{i}^{Q} ] $ to denote the class composition of client $ i $, where $ Q $ is the total number of classes. We further use vector $  V =[{\textstyle \sum_{i}^{}}N _{i}^{1},..., {\textstyle \sum_{i}^{}}N _{i}^{Q} ] $ to denote the composition of global dataset. Hence, the mismatch imbalance between local client $i$ and global dataset can be represented as the cosine similarity (CS) between their composition vectors, i.e., $ CS_{i} = (v_{i} \cdot  V ) / \parallel v_{i}\parallel  \parallel V\parallel  $. The mismatch imbalance can also be defined using other distance metrics such as Earth Mover's Distance \cite{zhang2021dubhe}.
\end{itemize}

Different types of class imbalance can result in different challenges in FL. The local imbalance may lead to poor classification performance of local models, which further harms the performance of the global model. Therefore, it is critical to tackle the local imbalance first to ensure a better global model \cite{ran2021dynamic}. 

Even if there is no local imbalance, the global imbalance of training data can also lead to performance deterioration, which has been mathematically proven in  \cite{duan2019astraea} \cite{zhang2021dubhe}. It has been pointed out that the global model is likely to converge at local optimum points when trained on a imbalanced global dataset. The performance degradation in FL can also be caused by the discrepancy between the distributions of the training set and the test set \cite{duan2019astraea}.

The mismatch imbalance is also detrimental to model performance in FL \cite{wang2020optimizing}. The variance in class distributions throughout the different training steps in FL will cause fluctuation in the model optimization \cite{zhang2021dubhe}. Moreover, when the data distribution among clients has less diversity, it will result in less diversity in generated local models. Thus, this will lead to the generated global model with poor robustness \cite{zhang2021dubhe}.

Combating the problem of class imbalance in FL is more challenging than that in traditional ML. This is because in FL, the data distributions of client datasets are usually confidential in order to protect user privacy. Moreover, the above three kinds of class imbalance are often intertwined together, making it crucial to develop techniques that joint handle class imbalance of different types \cite{li2019convergence} \cite{wang2020tackling} \cite{shen2021agnostic}.
\vspace{-0.2cm}

\section{Class Distribution Estimation}
\label{sec:class_estimation}
As discussed in Section~\ref{sec:preliminaries}, class imbalance in FL is more complicated than that in classical ML models as it may occur at local level, global level, and when there is discrepancy between local datasets. Thus, obtaining prior knowledge about class distributions is usually essential for designing solutions accordingly \cite{chakraborty2022improving}. Many existing approaches for tackling class imbalance of FL, such as cost-sensitive learning \cite{lin2017focal} \cite{wang2016training} \cite{khan2017cost} \cite{wang2018predicting}, need to know the data distribution.

However, estimating the actual class distribution in the FL setting is usually difficult. Although understanding the class distribution of a local dataset is rather straightforward, getting a full picture of the class distribution at the global level is non-trivial. This is because in the FL setting, only trained local models can be shared amongst client devices. Local datasets, however, must not be shared amongst clients to preserve user privacy. 

Therefore, many methods have been proposed to estimate global class distribution. We divide existing methods in the literature for estimating global class imbalance into two groups. The first group requires each client device to upload class distribution (i.e., not data itself) to the central server, which can be used to derive the global class distribution. The second group of methods infers class distribution based on trained models or uploaded gradients. In this section, we review the two groups of methods respectively.
\vspace{-0.2cm}

\subsection{Distribution Derivation Based on Local Distribution}
A straightforward method to estimate the global data distribution is to require clients to upload the class distribution of its local datasets to the server. The uploaded class distributions from all clients can then be merged into the global class distribution.

Mhaisen et al. \cite{mhaisen2021optimal} proposed an optimal user-edge assignment in the hierarchical FL method. The data distribution on the client side is required to be uploaded to the central server. 
Duan et al. \cite{duan2019astraea} \cite{duan2020self} proposed Astraea method, in which the clients send their local data distribution information to the server.

Requiring clients to upload data distributions is simple to operate. However, it has been pointed out that disclosing class distribution may still reveal part of the users' private information \cite{wang2019eavesdrop} \cite{fu2021cic}. Take FL for recommender systems as an example. The disclosure of label distributions of client data may reveal users' preferences on the client side. Although this method may compromise user privacy slightly, the core FL requirement of protecting clients’ raw data is still met \cite{mhaisen2021optimal}.
\vspace{-0.2cm} 

\subsection{Distribution Estimation Based On Model Parameters}
The second group of methods infers class distributions indirectly by examining the  models that are shared by clients. As such, these methods do not require clients to reveal class distributions directly, and hence, provides better protection of user privacy. Several types of model parameters, such as gradients, losses, and predictions, can be used to estimate data distribution of client devices. Next, we review existing methods that belongs to this group.

\textbf{Gradient-based methods.} Firstly, the relationship between the model's gradients and the sample numbers of different classes is analyzed. The model gradients are calculated on the class-balanced auxiliary datasets. And then the sample numbers of different classes can be estimated by the gradients.

However, this method needs auxiliary data which may be not available under certain conditions.

Wang et al. \cite{wang2021addressing} proposed a model for continuously monitoring the composition of training data when new data is constantly generated by client devices. It uses an auxiliary dataset that is balanced in class distribution. At time step $t+1$, the model downloads the global model $G_t$, feeds the samples in the auxiliary dataset into $G_t$, and obtains the gradient updates w.r.t. the auxiliary dataset. By comparing these gradient updates with the global model $G_{t+1}$ at time step $t+1$, the model can infer the global class distribution. Experimental results on real-world datasets show that the cosine similarity score between estimated and the ground-truth distribution achieves 0.98 on average. 

Yang et al. \cite{yang2021federated} revealed the correlation between gradients and class distributions: the expectations of gradient square of a trained model for different classes is approximately equal to the cardinality square of these classes. Based on this finding, they feed an auxiliary dataset into the trained global model to obtain the gradient updates for different classes. These gradient updates are then used to approximate the global class distribution of training data. 

Chen et al. \cite{chen2021novel} proposed a method for class distribution estimation based on Yang et al.' finding \cite{yang2021federated} about the correlation between gradients and class distributions. However, their method does not require auxiliary datasets. For each client, they compute the gradient updates of the client's trained model, from which they can directly derive the class distribution of the client's local dataset. The global distribution can then be formulated as the weighted average of all clients' class distributions, where the weights are set of parameters to be learned. Therefore, training the global model can be regarded as the process of minimizing the overall loss by quasi two-party theory \cite{mohri2019agnostic}. They propose an algorithm that optimizes the weights and model parameters alternatively for model training.  

Dong et al. \cite{dong2022federated} assumed that the data imbalance in FL may vary over time on both the local and global level. Moreover, they assume that there could be new data for unseen classes appear in the dataset. As such, a local or global model may forget information about old classes due to limited storage. To handle such a catastrophic forgetting problem, they propose a global-local forgetting compensation model. In particular, local clients will address the forgetting problem through a class-aware gradient compensation loss and a class-semantic distillation loss. On the global level, the model saves a set of old global models. Once an unseen class is encountered, the model uses a prototype gradient-based communication mechanism to send perturbed prototype samples of the new class to the proxy server. The proxy server reconstructs the perturbed prototype samples after receiving these gradients, and uses them to track the effectiveness of the saved old global models. The best performed old model will be used for relation distillation in order to tackle the global forgetting problem.

\textbf{Loss}: Usually, loss functions are used to guide the training process of deep learning models. The values of losses also reflect the performance of a model, i.e., higher loss indicates worse performance. There are studies that use techniques, such as active learning and reinforcement learning, to implicitly infer the data composition for optimizing global model.

For dataset with high data imbalance, the minority class usually have much higher loss than the majority class. In \cite{goetz2019active}, the value of loss function of one client is considered to reflect the usefulness of the client's dataset during a training round. Therefore, the clients that are considered more useful (i.e., having more samples belong to the global minority class) will be more likely to be selected in the subsequent training rounds in order to handle the issue of class imbalance. 

Shen et al. \cite{shen2021agnostic} added a constraint to the standard FL scheme: the empirical loss of every client should not overly exceed the average empirical loss. Under the heterogeneous data configuration with mismatch imbalance, this constraint is shown to make the global classifier to account for all classes equally, and hence, mitigate the detrimental effect of class imbalance. 

Chou et al \cite{chou2022grp} proposed a global-regularized personalization (GRP-FED) technique for FL. They aim at improving the fairness among all clients in terms of their contributions to the global model. If the standard deviation of the training loss in all clients is high, the training loss is quite different and the global model may suffer from client imbalance. 

\textbf{Performance of model}: The subsequent federated learning process is adjusted by the classification performance of the model on each class sample. The class with higher classification accuracy usually has a larger number of samples. This method cannot estimate the distribution information of the data, and can only guide the subsequent federated learning process through the classification performance of each class sample. For example, in client selection based methods for solving the calss imbalance problem, clients with better comprehensive classification performance are selected to participate in the training, so as to improve the performance of the global model.

To alleviate the impact of skewness in label distribution of clients' datasets, Mou et al. \cite{mou2021optimized} employed a balanced global validation dataset on the server side to score the performance of each client's model. This provides a straightforward, yet easy to operate method for client evaluation.

Geng et al. \cite{geng2022bearing} proposed to weight local models based on their F$_1$ scores in the model aggregation process. As such, the aggregation policy in FL is enhanced by increasing the weight of high-quality client models. 

Hao et al. \cite{hao2021towards} proposed the zero-shot data generation (ZSDG) algorithm, which can be used to generate labeled synthetic data for data-augmentation either at the client or the server side. ZSDG can be used in two ways, both of which can alleviate the data imbalance issue. First, ZSDG can utilize the pre-updated global model to generate synthetic data of the desired classes on the client side. Second, it can utilize the post-updated local models to generate synthetic data of the desired classes on the server side without access to any non-local data.
\vspace{-0.2cm} 

\subsection{Cluster}
This group of methods clusters clients into different groups based on the clients' certain attributes. In order to protect user privacy, the attributes disclosed by clients to the server should only be restricted to certain information such as model weights \cite{ghosh2019robust} \cite{xie2020multi}, gradients \cite{briggs2020federated} \cite{sattler2020clustered}, local optima \cite{ghosh2020efficient} \cite{mansour2020three}, etc. Based on the clustering results, the server can then use different strategies for different clusters in order to alleviate data balance issue, such as generating a personalized model for each cluster, data sampling and client selection, which aim to address global data imbalance. 

Zhao et al. \cite{zhao2020cluster} employed an agglomerative hierarchical clustering algorithm to cluster clients into groups. The features in clustering include the parameters of local models that are trained and uploaded by clients. 

Wang et al. \cite{wang2021adaptive} proposed a new weighted clustered FL (CFL) model based on adaptive clustering algorithm. At each round of FL training, clients are clustered into groups based on the cosine similarity of the uploaded local gradients. Each client not only uploads the local gradient but also uploads its class imbalance degree to the server.  

Fu et al. \cite{fu2021cic} proposed a class imbalance-aware clustered FL method (CIC-FL). Every client computes a feature vector, using weight updates and the label-wise gradients of the global model and sends it to the server. CIC-FL employs a top-down hierarchical clustering process. Then, the server iteratively conducts bi-partitioning to partition these clients into two clusters. 
\vspace{-0.2cm}

\subsection{Bottom-up Class Distribution Estimation}
Since clients have full access to their local models, they can infer the data distribution according to the state of the FL model, or infer the degree of similarity between the local data and the global data distribution.

To protect clients' privacy, Zhang et al. \cite{zhang2021fedsens} proposed a FedSens method, where clients decide whether to participate in this round of learning based on the current state, and the extrinsic-intrinsic reward. The current state that a client should consider is the classification performance on each class and the energy needed to train a local model at each device. The reward for each client depends on the client's dataset. If a client has balanced local data, the client should obtain a bigger reward.

In Dubhe \cite{zhang2021dubhe}, each participating client calculates the local data distribution, and then uses the homomorphic encryption method to encrypt the distribution. The server adds the distributions of all clients according to additive homomorphism, the result of which is then returned to all clients. The clients then decrypt the result to obtain the overall class distribution. 

Li et al. \cite{li2021sample} proposed a data selection method for FL. Instead of exposing local data or distribution of local data, each client only reports 1-bit information, i.e., if the client's data is relevant to the server or not. 

Hahn et al. \cite{hahn2019privacy} proposed an approximate Bayesian computation-based Gaussians Mixture Model called ``Federated ABCGMM'', which can oversample data in a minor class by estimating the posterior distribution of model parameters. The algorithm selects candidate parameters from the posterior distribution, and generate samples based on the selected parameters. These samples generated by the central server are then shared to clients, who will compute the similarity between these samples and their local samples. Based on the resulted similarities amongst all clients, the sampled parameter candidate will be determined to be accepted or rejected at the CS. When the process is iterated, the accepted set of parameters becomes close to its true posterior distribution.

These class estimation methods are listed in Table \ref{class imbalance estimation methods}.
\vspace{-0.3cm}

\begin{table*}[!t]
	\centering
	\caption{Representatives of reviewed class imbalance estimation methods in FL}	  
	\label{class imbalance estimation methods}     		       	
	\begin{tabular}{cp{6.5cm}p{6.5cm}}
		\hline
		Strategy & Representative articles& Detailed description \\ 
		\hline
		Updating distribution information& Mhaisen et al. \cite{mhaisen2021optimal}
		Geng et al. \cite{geng2022bearing}
		Mrad et al. \cite{mrad2021federated}
		Duan et al. \cite{duan2019astraea} \cite{duan2020self}& The server collects uploaded data distribution information, clients' privacy will be leaked to some extent.  \\ \hline
		Loss function& Shen et al. \cite{shen2021agnostic} Goetz et al. \cite{goetz2019active} Chou et al. \cite{chou2022grp}&The server estimates the class distribution according to the value of loss function\\  \hline
		Gradient& Chen et al. \cite{chen2021novel} Hao et al. \cite{hao2021towards} Wang et al. \cite{wang2021addressing} Yang et al. \cite{yang2021federated}&The server estimates the class distribution according to gradients\\ \hline
		Model performance& Mou et al. \cite{mou2021optimized}&The server estimates the class distribution according to model performance \\ \hline
		Clustering algorithm& Fu et al.\cite{fu2021cic} Zhao et al. \cite{zhao2020cluster} Wang et al. \cite{wang2021adaptive} & Clients are clustered into clusters, the server solves the class imbalance problem according to these clusters\\  	\hline	
		Bottom-up method&  Zhang et al. \cite{zhang2021fedsens} Zhang et al. \cite{zhang2021dubhe} Li et al. \cite{li2021sample}  Hahn et al. \cite{hahn2019privacy}&Clients estimate class distribution according to class distribution information obtained by typical protocol or model performance\\  		
		\hline
	\end{tabular}
	
\end{table*}
\vspace{-0.2cm}

\section{Classification with Class Imbalance in Federated Learning}
\label{sec:classification}
Class imbalance is common in real-world datasets. It may have a detrimental influence on how well classification models work due to the information gaps brought by limited sample sizes of a minority class, the overlaps between classes, and small disjuncts within one class \cite{ali2013classification}. The degree of such influence depends on the imbalance level, concept complexity, and size of training data \cite{japkowicz2002class}. To this end, there exists extensive investigation into class imbalance problem in traditional ML \cite{guo2008class} \cite{ali2013classification}, which can be categorized into five types in general \cite{xiao2021experimental}: preprocessing methods \cite{lopez2013insight}, cost-sensitive learning \cite{krawczyk2014cost}, algorithm-centered methods \cite{singh2015survey}, ensemble learning \cite{galar2011review}, and hybrid methods \cite{liu2008exploratory}. Most of these methods assume the setting of traditional ML where training data is located in a centralized place. In the setting of FL, however, training data is distributed on client devices, which poses new challenges in handling class imbalance. Therefore, new approaches tailored for FL have been proposed to address the class imbalance problem. Next, we review these approaches in this section.

\begin{figure*}[!t]
\centering
\setlength{\abovecaptionskip}{-0.2cm}
\includegraphics[scale=0.6]{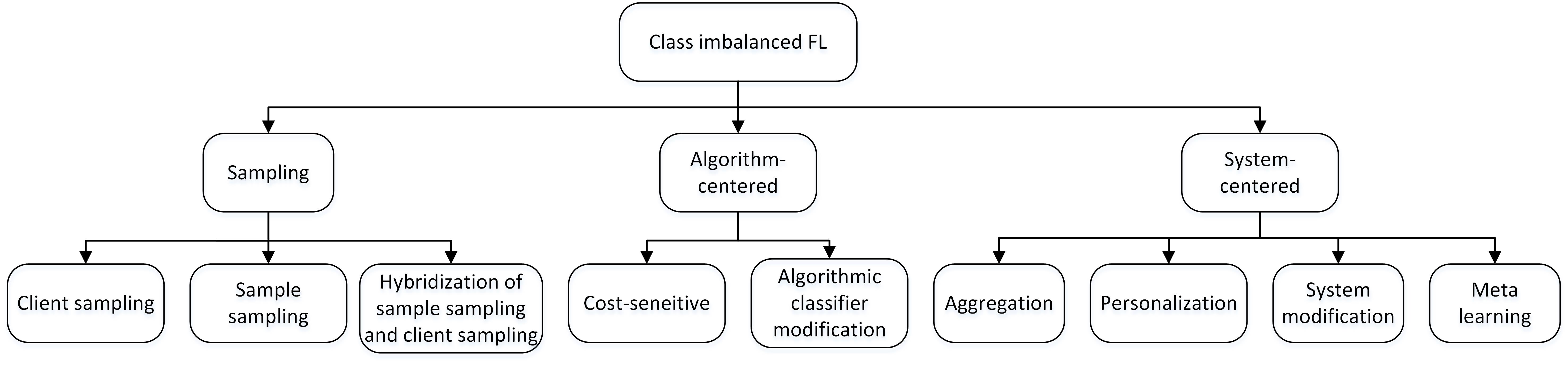}
\caption{Approaches for imbalance data.}
\label{Approaches for imbalance data}
\end{figure*}

Akin to the categorization schema used in traditional ML, we divide approaches for classification with class imbalance in FL into three categories: (1) sampling-based techniques; (2) algorithm-centered techniques; and (3) system-centered techniques, as shown in Fig. \ref{Approaches for imbalance data}. The sampling-based techniques can be seen as a sub-type of the preprocessing methods, which apply preprocessing steps to the training data to balance the class distribution. The algorithm-centered techniques aim to handle class imbalance by modifying the classification algorithm (e.g., changing the loss function and modifying the algorithm) and allowing the algorithm to focus more on the minority classes. The system-centered techniques address the class imbalance systematically in the FL framework, including methods that are based on aggregation, personalization, system modification, and meta-learning. 
\vspace{-0.2cm}

\subsection{Sampling Techniques}
Sampling based techniques selectively choose data that should be involved in the training process to generate a new sample space with balanced class distribution. In general, sampling based techniques are easier to operate compared with other groups of methods (e.g., algorithm-centered techniques) and do not require extensive expertise in ML. Thus, these methods have become popular when the models are operated by out-of-domain experts. Another strength of sampling based techniques is that they are agnostic to the core classification models  \cite{lopez2013insight}. As such, these techniques can be incorporated into any classification model, and event an ensemble of multiple classification models. 

However, straightforward data sampling is not guaranteed to boost the classification accuracy, and sometimes may even worsen the model performance on one client \cite{tangdata}. It is found that data sampling may generate local models that over-fit (under-fit) for the minority (majority) classes \cite{cao2019learning}. A possible explanation could be that naive data resampling may prevent clients from learning the ``Special Knowledge'' from local dataset in FL, which eventually leads to low accuracy \cite{cao2019learning}. Therefore, sampling strategies need to be carefully designed for FL in order to handle class imbalance effectively.

Therefore, many attempts have been made to propose sampling strategies that address data imbalance systematically in FL. In general, sampling can be done on either the data instance level (i.e., determining if one data instance should be involved in training) or the client level (i.e., determining if the data instances owned by one client should be involved in training). In this section, we categorize sampling techniques in FL into three groups and introduce them respectively: (1) data sampling; (2) client sampling; and (3) hybrid data and client sampling.
\vspace{-0.2cm}

\subsubsection{Data Sampling}
Given an imbalanced datasets, data sampling aims to balance the sample space with a pre-defined sampling strategy. Existing sampling strategies can be categorized into three groups \cite{haixiang2017learning}:

\begin{figure}[htbp]
	\centering
		\setlength{\abovecaptionskip}{-0.2cm}
	\includegraphics[scale=0.7]{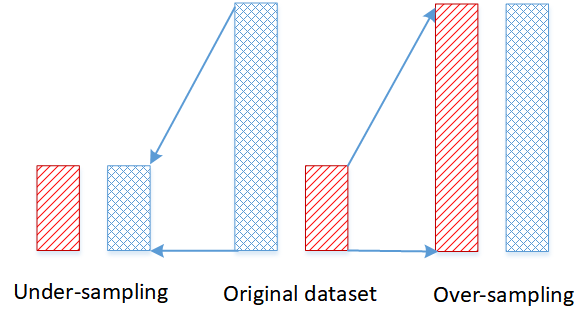}
	\caption{Sampling methods. }
	\label{Sampling methods}
\end{figure}

\begin{itemize}
\item  \textbf{Over-sampling:} generating new synthetic data instances for a minority class to increase its class size (see Fig. \ref{Sampling methods} ). Methods that belong to this group includes SMOTE \cite{chawla2002smote} and bootstrapping. 

\item \textbf{Under-sampling:} removing samples from the majority class to decrease its class size (see Fig. \ref{Sampling methods} ). A straightforward and yet efficient method is the random under-sampling (RUS) \cite{tahir2009multiple}, which randomly removes data from a majority class.

\item  \textbf{Hybrid sampling:} generating a balanced dataset via the combination of over-sampling and the under-sampling.
\end{itemize}

\textbf{Client-side data sampling.} In FL, each client can perform data sampling on its local datasets, i.e., oversampling the minority classes or undersampling the majority classes, to balance its local class distribution. Since the amount of local data owned by each client is usually limited in size, over-sampling is more preferable in the literature \cite{shingi2020federated} \cite{weinger2022enhancing}. Since the performances of local models are crucial to the final global model in FL, performing a quality local update with client side data resampling usually can significantly lead to a better model. 

\textbf{Server-side data sampling. } Solely performing data sampling locally on client side may not fully address the data imbalance issue systematically in FL \cite{duan2019astraea}. This is because clients may have data from different classes. Even if each client is balanced in class distribution, the aggregation of clients' datasets may not be balanced. Thus, server-side data sampling is required to address the issue. The server can choose a sampling strategy according to dataset characteristics that is estimated through the methods summarized in Section~\ref{sec:class_estimation}.

Shingi et al. \cite{shingi2020federated} employed SMOTE method to over-sample minority class at client to solve the local imbalance 

Hao et al. \cite{hao2021towards} proposed a novel zero-shot data generation (ZSDG) over-sampling method to mitigate class imbalance in the FL system. They studied two variants of ZSDG: client-side ZSDG and server-side ZSDG. In addition to handling class imbalance, ZSDG can also encourage more uniform accuracy performance (i.e., fairness) across clients in FL. 

Tijani et al. \cite{tijani2021federated} proposed a straightforward data extension method to deal with the severe label skew issue. Based on the classes that are missing from the client's private data, each client chooses a few extension samples from external data samples.

Weinger et al. \cite{weinger2022enhancing} examined the impact of data over-sampling on IoT anomaly detection performance in FL. They implemented a set of over-sampling techniques including random oversampling, SMOTE \cite{chawla2002smote}, and a variant of SMOTE known as ADASYN \cite{he2008adaptive}. They aim to improve the quality of the client's local model, ensuring that these clients can contribute meaningful updates to the global model. Thus, they over-sample data locally on the client side. 

Tang et al. \cite{tangdata} proposed an Imbalanced Weight Decay Sampling (IWDS) method, which is a simple but effective data resampling strategy. It dynamically regulates the sampling probability of different labels, accelerating the training process. In IWDS, the sampling weights of all data samples decay with the training rounds. In the early training rounds, in order to converge faster, it makes all clients to have more similar label sampling probabilities. In later rounds, to allow each client to better learn the special knowledge from its local datasets, it makes all clients to use their the original label sampling probability, which is a similar to local sampling.

To over-sample data in a minority class, Hahn et al. \cite{hahn2019privacy} presented an approximate Bayesian-based Gaussian mixture model that estimates the posterior distribution of model parameters across institutions while maintaining anonymity. At the server, plausible perturbed samples in the minority class are created, and when these samples are transmitted to the local client, they can improve the classification accuracy for the imbalanced data at the client's side.
\vspace{-0.25cm}

\subsubsection{Client Sampling}

\begin{figure}[h]
	\centering
	\setlength{\abovecaptionskip}{-0.2cm}
	\includegraphics[scale=0.45]{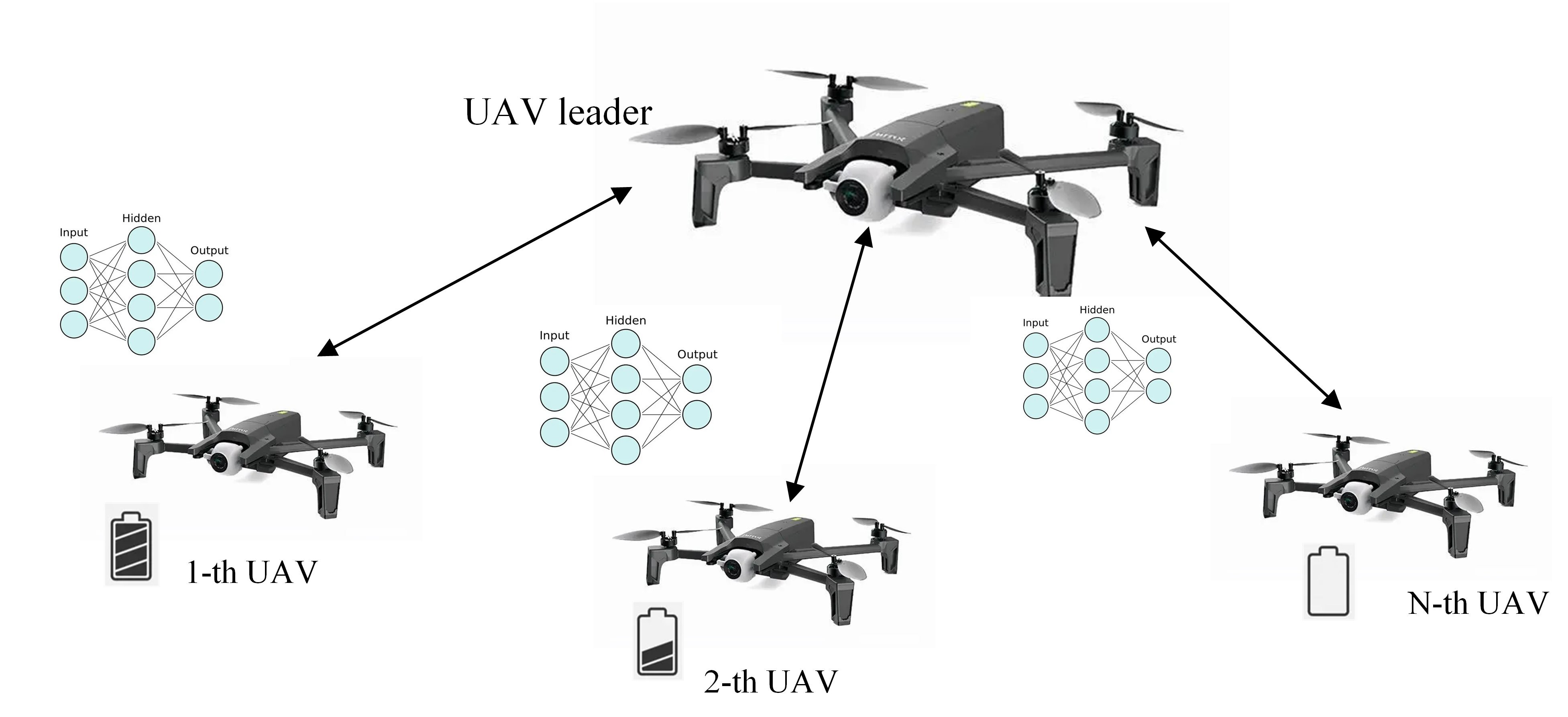}
	\caption{Client sampling.}
	\label{Client sampling}
\end{figure}
\vspace{-0.2cm}

Client sampling methods address the class imbalance issue by selectively choose clients that participate in a training iteration of FL. Such methods are especially useful when participating clients in FL are not always accessible, e.g., some clients may not have enough power or may be turned off when the server is polling. Fig. \ref{Client sampling} illustrates such an example of an unmanned aerial vehicles (UAV) network, where the participating vehicles are energy-constrained. Another reason for performing client sampling is that not all clients have access to data distributions that are comparable to the underlying global data distributions. The learning process would converge slowly or even diverge due to the lack of similarity between the distributions of specific clients and the global distribution \cite{chen2021novel} \cite{goetz2019active}. 

Random client selection may aggravate the data bias when data among clients are non-IID and the global data distribution is biased \cite{zhang2021dubhe}. As such, many methods select a client set with minimal class imbalance. However, this will result in less diversity in the local models trained at the clients' side. This may also result in a global model with poor robustness. 

Therefore, more sophisticated client sampling methods for FL have been proposed to resolve class imbalance issue. In this section, we review these methods. We categorize these methods by their sampling strategies: (1) optimization-based; (2) ranking-based; (3) clustering-based; and (4) bottom-up based. 

\textbf{Optimization-based client sampling.} This group of methods formulates the client selection as an optimization problem which maximizes a certain objective (e.g., overall model performance) given certain constraints (e.g., communication cost). 

Mrad et al. \cite{mrad2021federated} proposed a solution to the problem of class imbalance in FL for energy-constrained UAV networks while considering the limited availability of UAVs due to their stringent energy constraints. Their aim is to obtain a reliable and stable performance of FL for UAVs network, while considering the class imbalance problem. The optimization strategy is to select the set of UAVs that delivers the lowest class imbalance, to improve classification accuracy.

Mhaisen et al. \cite{mhaisen2021optimal} studied an User-Edge-Cloud hierarchical FL paradigm. In order to reduce the distribution difference between edge devices, they formulate the client sampling as a problem which assigns users (clients) to edge nodes within their communication range, which is can be viewed as an NP-Hard problem. They suggested two solutions to this problem. The first one is a branch-and-bound based solution to a simplified linear version of the problem that assumes equal assignments between edge devices. The second solution uses heuristics to greedily balance out class distributions among edge devices. 

Yang et al. \cite{yang2021federated} employed combinatorial multi-Armed bandit (CMAB) \cite{chen2013combinatorial} to design a client set selection method based on a composition vector. This method can select a client set with minimal class imbalance. 

\textbf{Ranking-based client sampling.} Clients can be evaluated and ranked by indicators such as local data distribution and local model classification performance. Ranking-based client sampling sort the clients according to a chosen indicator, and select clients based on the ranking result.

Chen et al. \cite{chen2021novel} presented a client sampling method which assigns a reward for each client. The reward of a client is associated with the cosine similarity between the client's and the global  class distribution. Thereafter, all clients are ranked by their rewards and only the top ranked clients are allowed to participate in the global model updating.  

In \cite{goetz2019active}, an active FL model is proposed. In each FL training round, it selects an optimized subset of clients based on a value function that reflects how useful the data of that client is. They propose that a natural value function that can be leveraged is the local model's loss value: samples of the minority class will have significantly higher loss value than majority class samples. Thereafter, server converts all client's value scores into the probability of each client being selected in the next round. As a result, in the next training round, higher-valued clients will have higher probabilities of being selected, which implicitly favors those clients with more samples for minority classes.

\textbf{Cluster-based client sampling.} This group of methods groups clients into clusters and select client based on their associated clusters to ensure the class balance of the global dataset.

Zhao et al. \cite{zhao2020cluster} proposed a cluster-based solution to improve the accuracy of the minority class at the low expense of performance degradation on the majority class. Firstly, it performs FedAvg. Then, when the model is nearly stable, it clusters clients into groups using the hierarchical clustering method. Finally, participants are selected from these clusters for the current round of training.

\textbf{Bottom-up client sampling.} In this group, the participation of a client is determined by the client itself (bottom) instead of centrally by the server. Clients observe the data distribution of the global dataset or the performance of the current global model, and then determine to participate into the global training or not. 

In \cite{zhang2021dubhe}, the data distribution information of each client is encoded using homomorphic encryption. The server adds the encrypted distribution information and then returns it back to clients. Each client calculates a participation probability according to this class distribution. The method also proposed a multi-time client selection method to further balance the global dataset in each training round. 

Zhang et al. \cite{zhang2021fedsens} proposed a bottom-up device selection method to solve the class imbalance problem in abnormal health detection applications. To protect client's private, the clients judge whether to participate in this round of learning based on current state (the classification performance on each class and the energy needed to train a local model at each device) and extrinsic-intrinsic reward. The extrinsic reward is a feedback provided by current environment, it is defined as the global model's performance. The intrinsic reward considers the significance of the local model update and the energy cost of the device. 
\vspace{-0.3cm}

\subsubsection{Hybrid Data and Client Sampling}
The strategies of data sampling and client sampling can be further combined to achieve more improvement in model performance. In particular, client selection is often combined with global data sampling to solve the global imbalance problem in FL. 

Duan et al. \cite{duan2019astraea}\cite{duan2020self} proposed a self-balancing FL framework named Astraea. In \cite{duan2019astraea}, they employ data augmentation (over-sampling) method, to solve the global imbalance problem. In \cite{duan2020self}, it firstly performs both the z-score-based data augmentation and the under-sampling to solve the global imbalance of training data, according to distribution information from clients. Then, it uses a mediator which asynchronously receives and applies these updated models from participated clients to average the local imbalance. Therefore, the mediator can obtain a more balanced model by rescheduling clients' training. 

Li et al. \cite{li2021sample} presented a productive hierarchical sample selection system that first chooses the best clients, and then their top-notch samples. Before training, they select clients relevant to the target FL task using a private set intersection based approach. Using a determinantal point process based method, they maximize both statistical homogeneity and content diversity of these chosen clients within one budget. Then, a selection strategy based on erroneous-aware importance is proposed to dynamically choose significant clients and samples during each iteration in FL training process. 
\vspace{-0.3cm}

\subsection{Algorithm-Centered Techniques}
Unlike sampling methods, which only apply pre-processing steps to data, algorithm-centered methods make changes to the classification algorithm to handle class imbalance. This includes changing the loss function and modifying the classification algorithm itself, which aim to make the classification algorithm more sensitive to minority classes.
\vspace{-0.2cm}

\subsubsection{Cost-Sensitive Learning}
The classical loss function used for classification task, e.g., cross-entropy, assumes all data samples are independent, and applies same penalty to misclassifcation errors for all data samples. This makes it prone to class imbalance, when samples do not confirm IID distribution. Thus, cost-sensitive learning has been proposed, which enhances traditional loss classification loss function via penalizing a model more if the model misclassifies a sample from the minority class, i.e., assigning higher misclassification costs for the minority class. 

In general, cost-sensitive learning approaches can be seen as re-weighting strategy that weigh data samples in the computation of loss function according to their importance. As such, it shares similar spirit with data resampling, except that cost-sensitive learning is a soft reweighting strategy, while data resampling assigns hard (integral) weights to data samples. Compared with data over-sampling, cost-sensitive learning does not increase dataset size. This offers more efficient training of model, making it potentially more preferrable for massive data streams.However, compared with resampling techniques, cost-sensitive approaches seem less popular. There are two possible reasons \cite{krawczyk2014cost}. First, it is challenging to define the proper cost values in these approaches. Most of the time, the optimum cost values cannot be determined solely by observing the data. As such, an expensive finetuning process needs to be performed to search for the best cost values that optimize the model performance. Second, cost-sensitive learning often needs to modify learning algorithms \cite{haixiang2017learning}, and thus, it requires developers of FL applications of FL to have solid understanding of the classification algorithm.

Lin et al. \cite{lin2017focal} introduce the focal loss which is an improved version of the traditional cross entropy (CE) loss for binary classification. For one data instance, CE can be formulated as:
\begin{subequations}\label{eqn-3}
\begin{numcases}{CE\ (p,y)=}
-\log(p)& \text{if} y=1 \label{eqn-3-1}\\
1-\log(p)& otherwise \label{eqn-3-2}
\end{numcases}
\end{subequations}
where $y \in  \left\lbrace \pm 1 \right\rbrace  $ represents the ground-truth class of the data instance, and $ p \in [0,1] $ represents the model’s predicted probability of the instance being labelled as $1$. For ease of presentation, we define the probability  $p_{t}$ of the instance being classified as the ground-truth label:
\begin{subequations}\label{eqn-4}
\begin{numcases}{p_{t}=}
p& \text{if} y=1 \label{eqn-4-1}\\
1-p& otherwise \label{eqn-4-2}
\end{numcases}
\end{subequations}
Thus, CE can be simplied as $ CE\ (p,y) = -\log(p_{t}) $.

Lin et al. reshape the loss function to down-weight easy positive samples (i.e., majority class samples) and thus concentrate on learning the more difficult samples around decision boundaries (i.e., minority class samples). They introduce a coefficient to achieve this. The focal loss (FL) is formulated as:
\begin{equation}
FL\ (p_{t})= -(1 - p_t) ^{\gamma} \log(p_{t})
\end{equation}
where $\gamma$ is a hyperparameter that controls how much the model should focus on the more difficult samples. When $\gamma =0$, FL is equivalent to CE.

There are two properties of $FL$: (1) The coefficient $1-p_t$ is close to 1 when the value of $ p_{t} $  is small (i.e., the sample is severely misclassified). The factor decreases to 0 if $p_{t}$ is close to 1 (i.e., the sample is correctly classified with high confidence); (2) The rate at which easy classified samples are down-weighted is smoothly adjusted by the focusing parameter $\gamma$.

Wang et al. \cite{wang2021addressing} proposed a monitoring scheme that can infer the distribution of global training data for each FL round according to gradients. They design a new Ratio Loss to mitigate the impact of the imbalance. Once the monitor detects a similar imbalanced distribution continuously, it will acknowledge clients to apply a mitigation strategy that is based on the Ratio Loss function. 

Sarkar et al. \cite{sarkar2020fed} proposed a new loss function called Fed-Focal Loss. It reshapes CE during training. According to the client's training performance, it down-weights the loss assigned to well-classified examples and focuses on those harder examples. Moreover, by leveraging a tunable sampling framework, the server selects the best-performing clients for global training. Therefore, this method improves the robustness of the global model. 

Wang et al. \cite{wang2021federated} introduced a modified CE loss function, balanced cross entropy (BCE). Moreover, they define a structural loss function, which is designed to prevent overfitting. Then a parameter $ \lambda $  is to balance these two losses.

Shen et al. \cite{shen2021agnostic} impose constraints on the standard FL formulation so that the empirical loss of every client should not overly exceed the average empirical loss. Such constraints are shown to force the classifier to account for all classes equally and hence mitigate the detrimental effect of class imbalance, under a type of heterogeneous data configuration that captures the mismatch between the local and global imbalance. Their formulation can significantly improve the testing accuracy of the minority class, without compromising the overall performance.

A Global-Local Forgetting Compensation model was proposed by Dong et al. \cite{dong2022federated}. It mainly comprises of a class-aware gradient compensation loss and a class-semantic relation distillation loss to battle local catastrophic forgetting induced by the class imbalance on the client side and global catastrophic forgetting caused by non-i.i.d. class imbalance across clients. While using the old global model with the best performance, a prototype gradient-based communication mechanism is created between the proxy server and clients for their private communication.
\vspace{-0.2cm}

\subsubsection{Algorithmic Classifier Modification}
\begin{figure}[!t]
	\centering 	
	\subfloat[Original decision boundary.]{\includegraphics[scale=0.52]{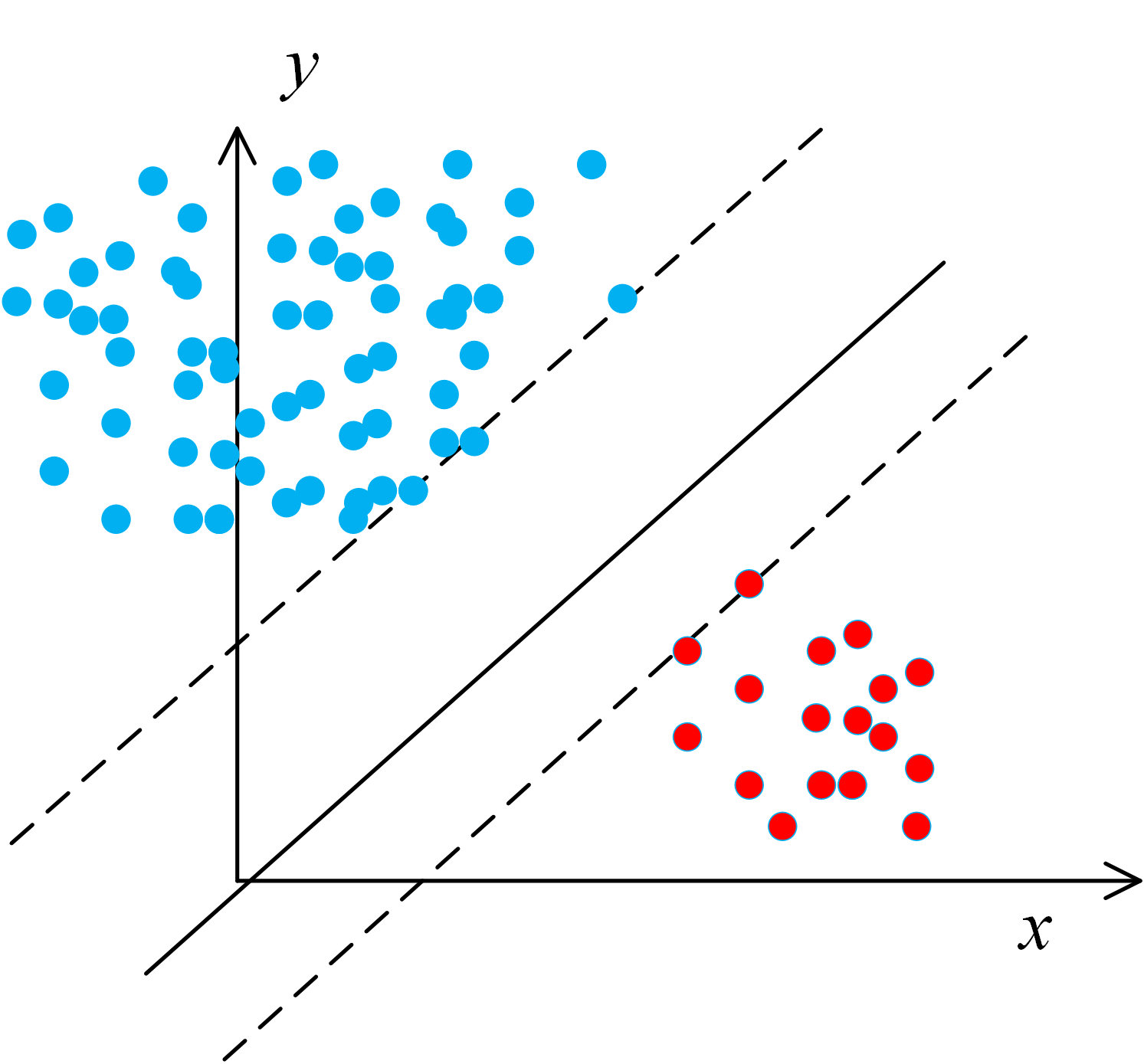}
		\label{algorithm1}}
	\hfil
	\subfloat[Dynamic decision boundary.]{\includegraphics[scale=0.52]{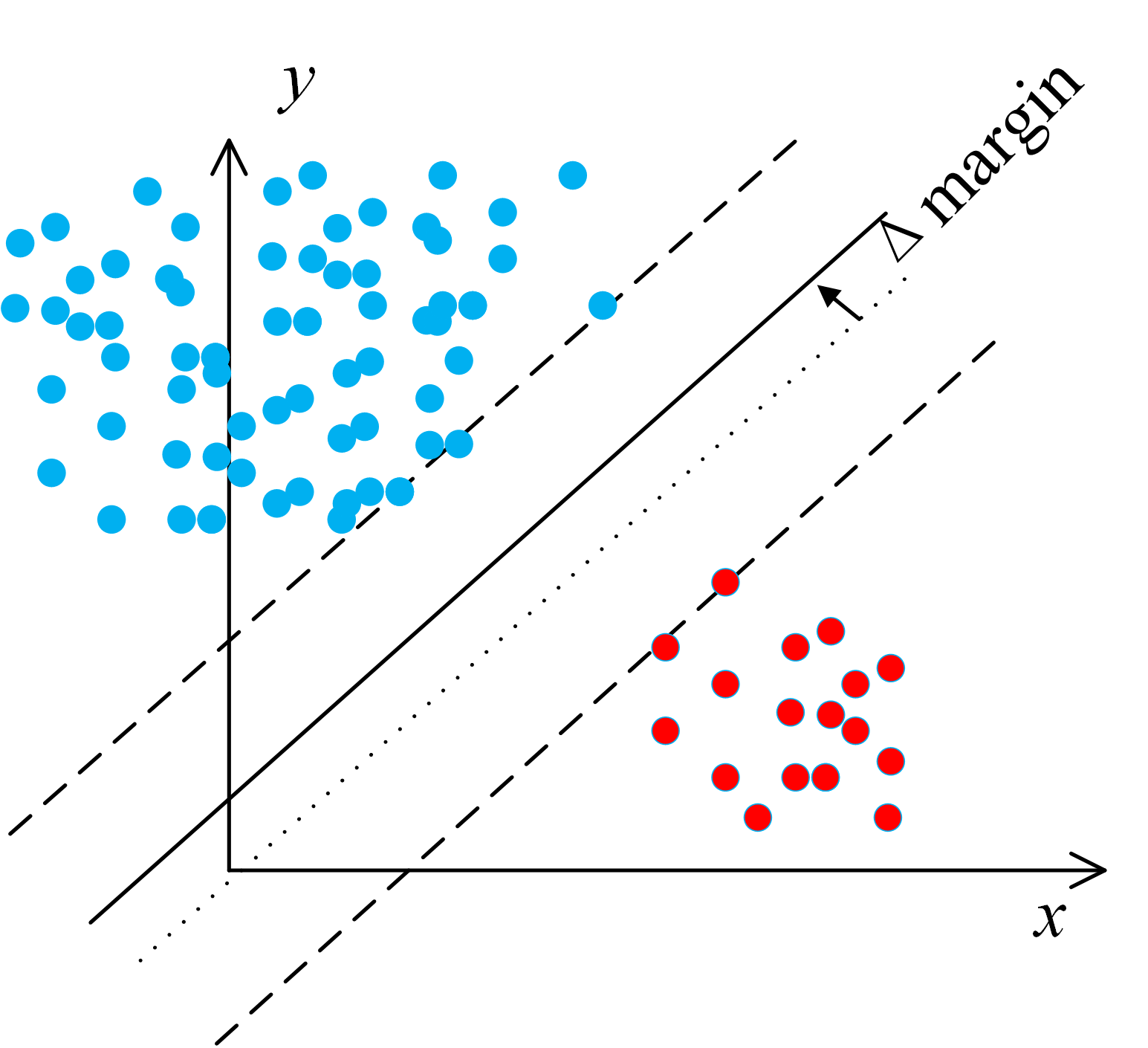}
		\label{algorithm2}}
	\caption{Algorithm-centered approach.}
	\label{algorithm}
\end{figure}

Some algorithm-centered methods directly modify the learning process to increase the classifier's sensitivity to minority classes \cite{khan2017cost}. As shown in Fig. \ref{algorithm}, the method encourages a classification model to have a larger margin between a minority class and other classes. Taking dynamic margin softmax as an example, it adds $ \Delta $margin for minority class.

Ran et al. \cite{ran2021dynamic} presented a method which creates dynamic margins in FL. The classifier which is trained using class imbalanced data, it's original decision boundary tends to be closer to minority classes. It encourages each participating client's local classification model to have large margins for minority classes, a dynamic term is skillfully added to the loss function which guide the model to learn more separable features from data.  Accordingly, in the local training phase, the original softmax is replaced with the dynamic margin softmax. The CE loss is further combined with the dynamic margin approach.

Hua et al. \cite{hua2020blockchain} improved the traditional SVM model by giving the majority and minority classes different penalty components. This technique is capable of handling data with imbalanced traction and braking. 

Federated fuzzy learning approach with class imbalanced data was proposed by Dust et al. \cite{dust2021federated}. The main idea is to virtually oversample the minority class samples.  They presented an imbalance adaption mechanism to increase the impacts of minority class samples in the fuzzy learning process in order to address the problem of class imbalance. By allowing the competent rules of the minority class to coexist with those of the majority class, this has the effect of preventing the model from being biased against the majority class. 

Li et al. proposed  \cite{li2021fedrs} a FedRS method. During local procedures, they advocate that the update of missing classes’ proxies should be restricted, and thus, a restricted softmax is introduced. Adding ``scaling factors'' to the softmax operation is a simple method to put this into practice. Then, a fine-grained aggregation is based on $p_{k,c}=\frac{N_{k,c}}{ {\textstyle \sum_{k}^{}} N_{k,c}}$, where $N_{k,c}$ is the number of $ c $-th class samples on $ k $-th client. 
\vspace{-0.2cm}

\subsection{System-Centered Techniques}
System-centered methods address class imbalance by changing the structure of FL framework. We further divide them into aggregation-based method, personalization method, and system modification method.
\vspace{-0.2cm}

\subsubsection{Aggregation Method}
In the model aggregation stage, the weight of each local model can be adjusted according to a certain evaluation metric. The standard FedAvg, which assumes class balanced training data, only considers data volumes of clients when weighing them in aggregation. However, the weights of clients can also be manipulated for the purpose of handling class imbalance issue.

Wang et al. \cite{wang2021adaptive} proposed a new weighted clustered FL (CFL) model based on an adaptive clustering algorithm. Weighted per-cluster model aggregation is performed on the server side. Each cluster is assigned with a different weight to balance the contribution of each class of global training data. In addition, the weight of each cluster is optimized through the convergence rate analysis. 

Geng et al. \cite{geng2022bearing} proposed a weighted aggregation policy based on F$_1$-scores. To tackle class imbalance, local models of clients are weighted based on the weighted F$_1$-scores, which implicitly handles the class imbalance issue. They also proposed an improved aggregation algorithm that is based on the accuracy difference between client temporary accuracy and one base accuracy. This method can not only improve the accuracy of model aggregation, but also reduce communication time. 

Mou et al. \cite{mou2021optimized} proposed to generate a global balanced dataset on the server side. Then a validation score $ s_{i} $ is computed by evaluating the performance of the local model of client $ i $ on the balanced global dataset. Finally, each client aggregation weight is computed based on the validation score. They tested different options for validation scores such as accuracy, IoU, and cross-entropy loss. 
\vspace{-0.2cm}

\subsubsection{Personalization Method}

\begin{figure*}[h]
\centering
\setlength{\abovecaptionskip}{-0.2cm}
\includegraphics[scale=0.5]{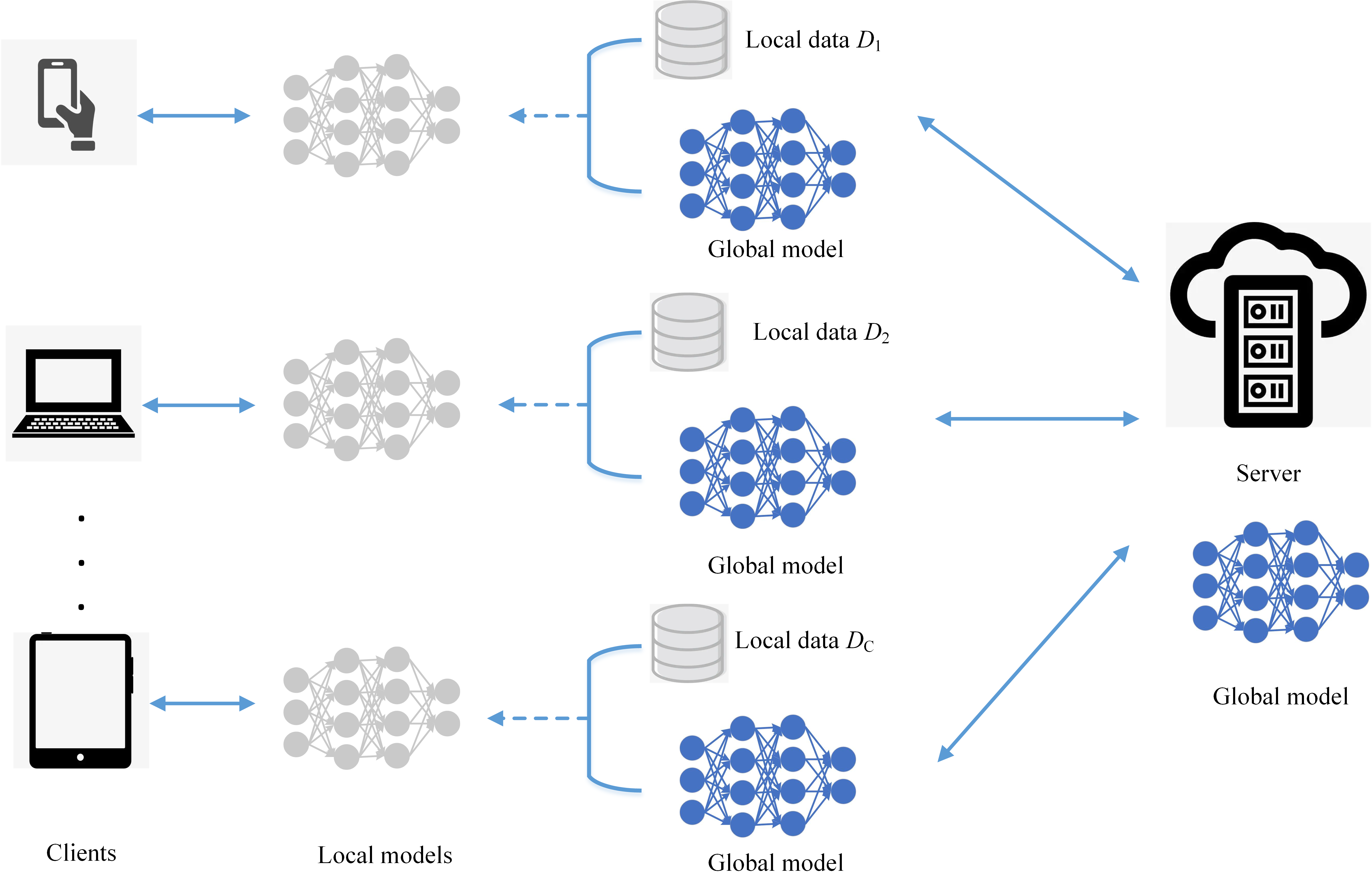}
\caption{An illustration of personalization FL. }
\label{personalization}
\end{figure*}

In traditional FedAvg algorithm, all clients are treated equally. As such, the differences between data distributions of clients are ignored, which leads to poor model performance in the case of high class imbalance. Further, as FedAvg weighs clients by their data volumes, those major clients can easily dominate the global model, while other clients are overlooked in the aggregation. Therefore, the classification performance drops significantly on them. Even with algorithms that encourage these overlooked clients to focus on global fairness \cite{mohri2019agnostic} \cite{li2019fair}, the performance gap between the global and local tests may remain significant \cite{jiang2019improving}. 

All these indicate that personalization is essential in the FL model \cite{chou2022grp}. With personalization, local clients put unique focus on their local data when training \cite{fallah2020personalized} \cite{khodak2019adaptive} \cite{liang2020think}. As shown in Fig. \ref{personalization}, for each client cluster or each client, a personalized local model is generated based on the global model and its own local data.

Fu et al. \cite{fu2021cic} proposed a CIC-FL method, where each client uses over-sample or under-sample to generate a balanced local dataset. Thereafter, clients are clustered into different groups. To address the class imbalance and data shift problem, one personalized model is trained in each cluster. 

Chou et al. \cite{chou2022grp} presented a Global-Regularized Personalization (GRP-FED) method into FL, which addresses the problem of class imbalance by taking into consideration a single global model and personalized local models for different clients. The global model mitigates the problem of the global long-tailedness and treats clients fairly by using adaptive aggregation. Each client's local model is learned from the local data and aligns with its distribution for personalization. GRP-FED uses an adversarial discriminator to regularize the learned global-local features, preventing the local model from simply overfitting. 

To process the contextual access abstraction for IoT devices, Yu et al. \cite{yu2020learning} designed a customized data augmentation approach. They design two mechanisms to address unbalanced records: context random sampling and adding contextual noise. The goal is to reduce the the correlation between environment variables/device states and IoT access with a certain action by performing random sampling on the environment variables and device states. Noise is added to the environment variables of the attack sample.

Chen et al.\cite{chen2022personalized} designed a Deputy-Enhanced Transfer to smoothly transfer global knowledge to clients' personalized local models. They proposed a Conjoint Prototype-Aligned (CPA) loss to address the class imbalance problem and make the FL framework's balanced optimization easier. The CPA loss computes the global conjoint objective based on global imbalance and then adjusts the client-side local training through the prototype-aligned refinement to eliminate the imbalance gap with such a balanced goal, taking into account the inaccessibility of clients' local data to other clients and the server in the FL model.
\vspace{-0.2cm}

\subsubsection{System Modification Method}
The class imbalance problem can also be addressed by revising the FL framework systematically.

Convolutional neural networks (CNNs) are used on both the client and server sides by Cheng et al. \cite{cheng2022blockchain} to extract practical features. The imbalanced input data causes the neural network to extract unbalanced features, so they first get a cluster for each class, which is then used to build the classifier. Then, using the categorical cross-entropy function as a guide, each client model is trained. With the help of blockchain technique, the traditional FL is enhanced without having to worry about the failure of the server and boosts the privacy of clients.

Cheng et al. \cite{cheng2022class} proposed a class-imbalanced heterogeneous FL method. They use a convolutional neural network (CNN) on both the client and server sides to extract useful features. On both the client side and the server side, a prototype is obtained by feature extraction for each class to the latent space. Then, a classifier is built based on the obtained prototypes. Therefore, the proposed method does not require information about the distribution of the raw data, and thus effectively preserves privacy.  

Giorgas et al. \cite{giorgas2020online} proposed an online FL with Sampling (OFLwS). The objective of online FL (OFL) \cite{yang2019federated} is to take advantage of the additional resources and training samples that are available either in a central node or in the edge nodes. OFL repeats the share-train-merge process several rounds. The proposed OFLwS builds on the OFL, and samples training data that is used in each retraining step. 

Chakraborty et al. \cite{chakraborty2022improving} proposed a two phases FL model. The first step is that the local models are pre-trained using an autoencoder. Then these local models are re-trained with deep learning model. They showed that the autoencoder pre-training is useful for the class imbalanced FL model. Moreover, they propose an adaptive focal loss for severely class imbalanced datasets in the FL model.
\vspace{-0.2cm}

\subsubsection{Meta Learning Method}
Meta learning, which means learning to learn, was born with the expectation of mimicking the ``learning ability'' of human beings. In ML, a large amount of data is used to train a model. However, when the scenario changes, the trained model may need to be retrained. For humans, however, we can easily adapt our pre-acquired skills to unseen scenarios. Meta learning is such a framework that aims to enable a ML model to obtain such capability: learning new tasks on the basis of existing ``knowledge''. Meta learning can also be used improve the classification performance of minority class in FL: adapting a model's existing knowledge of the majority class to the minority class \cite{zheng2021federated} \cite{zheng2020novel}.
\vspace{-0.3cm}

\begin{figure}[htbp]
\centering
\setlength{\abovecaptionskip}{-0.2cm}
\includegraphics[scale=0.6]{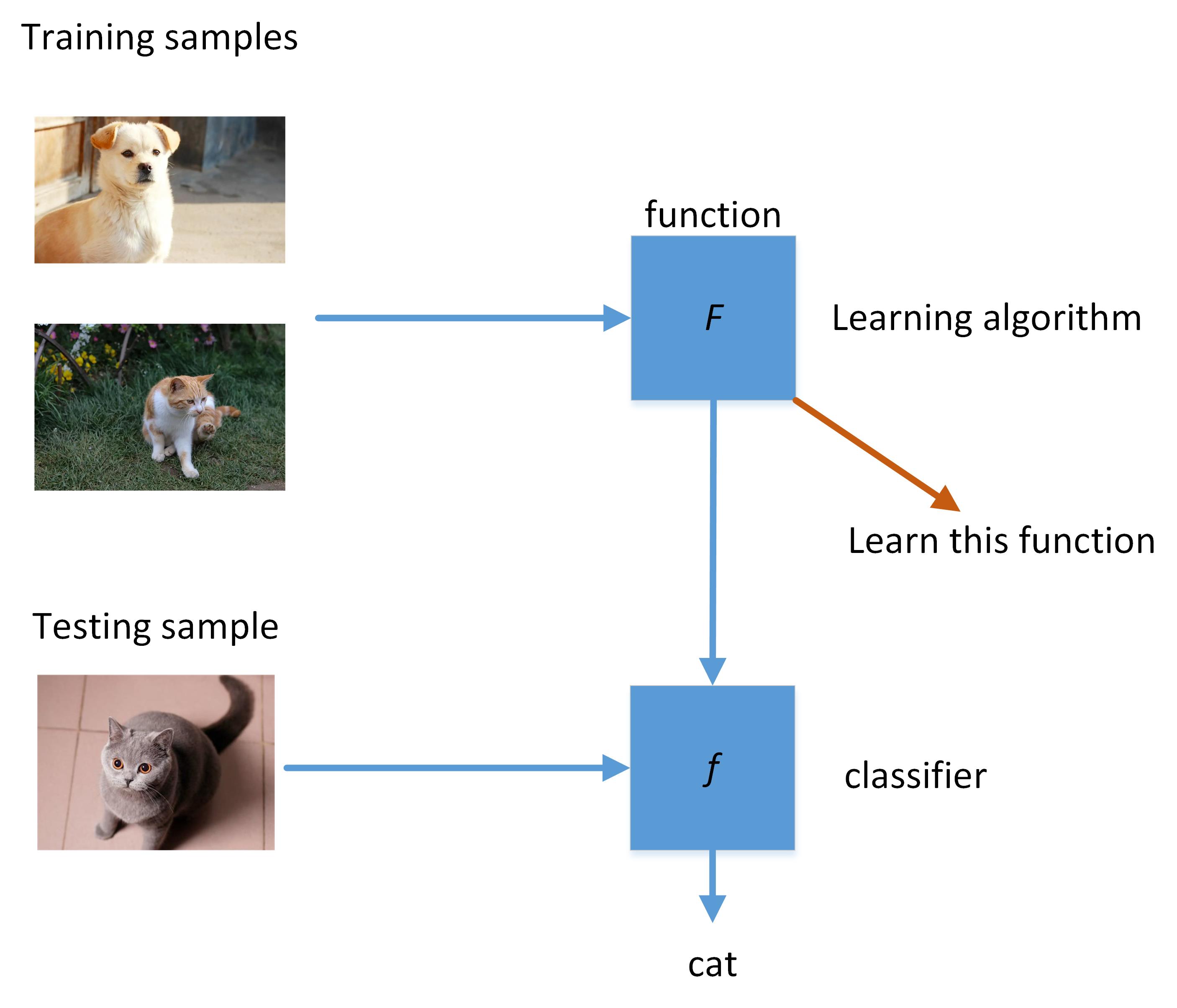}
\caption{An illustration of meta-learning.}
\label{meta-learning}
\end{figure}
\vspace{-0.2cm}

Fig. \ref{meta-learning} illustrates the key idea of meta-learning, which includes three steps: 
\begin{itemize}
\item Identify a function (e.g. linear regression model and a neural network) with a set of learnable parameters (i.e., variables).
\item Define a loss function with respect to these learnable parameters.
\item Find the optimal parameters to minimize the loss function.
\end{itemize}

The whole process above can be seen as a learning algorithm, and the goal of this process is to find a function with certain parameters. The learning algorithm itself can be regarded as a function, represented by $ F $, its input is training data, and the output is a classifier (a function). $ F $ is usually hand-crafted, so it is possible to learn this function directly, i.e., learning about how to ``learn''.

For the objective of detecting fraudulent credit cards, Zheng et al. \cite{zheng2021federated} proposed a meta-learning model that utilizes the FL technique. To accomplish few-shot classification, they propose an enhanced triplet-like metric learning method called the deep $ K $-tuplet network. To enable joint comparison with $ K $ negative samples in each mini-batch, this network specifically generalizes the triplet network. Additionally, they create a FL model based on a $ K $-tuplet network that can safeguard data privacy while allowing the FL model to be shared with several banks. 

These approaches that solve the class imbalance problem in FL are summarized in Table \ref{solving methods}.

\begin{table*}[!t]
\centering
\caption{Overview of literatures for class imbalanced FL}	   
\label{solving methods}    		       	
\begin{tabular}{cp{5.5cm}p{6cm}}
	\hline
	Strategy& Representative articles&Detailed description \\ 
	\hline
	Sample sampling & Shingi et al. \cite{shingi2020federated} Hao et al. \cite{hao2021towards} Tijani et al. \cite{tijani2021federated}  Weinger et al. \cite{weinger2022enhancing} Tang et al. \cite{tangdata} Hahn et al. \cite{hahn2019privacy}& Class balance is achieved by sampling samples from different classes \\ \hline
	
	Client sampling& Mrad et al. \cite{mrad2021federated} Mhaisen et al. \cite{mhaisen2021optimal} Yang et al. \cite{yang2021federated} Goetz et al. \cite{goetz2019active} Zhao et al. \cite{zhao2020cluster} Zhang et al. \cite{zhang2021dubhe} Zhang et al. \cite{zhang2021fedsens}&Class balance of the global dataset is achieved by client selection \\ \hline
	
	Hybridization sampling&Duan et al. \cite{duan2020self} \cite{duan2019astraea} Li et al. \cite{li2021sample}&Class balance of the global dataset is achieved by client selection and sample sampling\\ \hline
	
	Cost-sensitive learning& Wang et al. \cite{wang2021addressing} Sarkar et al. \cite{sarkar2020fed} Wang et al. \cite{wang2021federated}  Shen et al. \cite{shen2021agnostic} Dong et al. \cite{dong2022federated}&Modify learning objectives by assigning higher misclassification costs for minority class samples than for majority class samples\\ \hline
	
	Algorithmic classifier modification& Ran et al. \cite{ran2021dynamic} Hua et al. \cite{hua2020blockchain} Dust et al. \cite{dust2021federated}  Li et al. \cite{li2021fedrs} & It directly modifies the learning process to raise the classifier's sensitivity to minority classes\\ \hline
	
	Aggregation method& Wang et al. \cite{wang2021adaptive} Geng et al. \cite{geng2022bearing} Mou et al.  \cite{mou2021optimized} &Local models are weighted in model aggregation based on weight which may reflect the classification performance of local models\\  \hline
	
	Personalization method&Fu et al. \cite{fu2021cic} Chou et al. \cite{chou2022grp} Yu et al. \cite{yu2020learning} Chen et al. \cite{chen2022personalized}&Personalized models are trained for each client or for each client cluster\\ \hline
	System modification method& Cheng et al. \cite{cheng2022blockchain} \cite{cheng2022class} Chen et al. \cite{chen2021novel} Giorgas et al. \cite{giorgas2020online} Chakraborty et al.\cite{chakraborty2022improving} &It changes the structure of the FL model \\ \hline
	Meta-learning method&Zheng et al. \cite{zheng2021federated}&It use meta-learning to improve the classification ability of minority class \\ 
	\hline
\end{tabular}
\end{table*}
\vspace{-0.2cm}

\subsection{Comparative Analysis}
In this section, we compare the characteristics of all the methods discussed above, which solve the class imbalance problem after the data distribution estimation is completed. Table \ref{analysis} presents the advantages and disadvantages of these methods.

\begin{table*}[!t]
\centering
\caption{Comparative analysis of the class imbalance solutions}	     
\label{analysis}  		       	
\begin{tabular}{cc p{6cm}}
	\hline
	Approach& Advantage&Disadvantage \\ 
	\hline
	Sample sampling&Simple, no algorithm knowledge is required&Changes of over-fitting in oversampling\\
	&Flexible, it can be applied to any classification task&Changes of information loss in undersampling\\ 
	&&Only local class imbalances can be resolved\\ \hline
	Client sampling&Flexible, it can be applied to any classification task&Server or clients need to evaluate the current environment and make the best decisions\\
	&No algorithm knowledge is required&\\ 
	&Clients can have the autonomy to participate& \\ \hline
	Hybridization sampling&Flexible, it can be applied to any classification task&The learning process is too complicated\\
	&No algorithm knowledge is required&Server or clients need to evaluate the current environment and make the best decisions\\ 
	&Performance improvement&\\
	\hline
	Cost-sensitive learning&Computational effective &The cost assessment process is difficult\\
	&&Insufficient when optimal cost values are missing\\ 
	&&Knowledge of algorithms is required\\
	\hline
	Algorithmic modification&Better performance&Algorithm explicit\\ 
	&&Majority class performance degradation\\ 
	&&Overdriving the classifer toward minority	class\\ 
	&&Knowledge of algorithms is required\\ \hline
	Aggregation method&Simple, no algorithm knowledge is required&Local models need to be evaluated\\ \hline
	Personalization method&Targeted&Cannot generate a global unified model\\
	&Better performance for each client &                             \\ \hline
	System modification method&Distribution estimation is not required&The learning process is too complicated\\
	&&The system should be adjusted according to the task characteristics\\ \hline
	Meta-learning method&Distribution estimation is not required&The learning process is too complicated\\ 
	&&Meta-learning knowledge is required\\
	\hline
\end{tabular}
\end{table*}
\vspace{-0.2cm}

\section{Federated Learning Performance Evaluation}
\label{sec:metric}
In this section, we first introduce the metrics used to evaluate classification performance in traditional ML. Then, we introduce the performance metrics that are used to evaluate the FL models considering the class imbalance issue. 
\vspace{-0.2cm}

\subsection{Model Evaluation Metrics for Classification Performance}
Model evaluation is critical for us to determine the performance of any ML models. Therefore, choosing proper and indicative evaluation metrics is of great importance for understanding a ML model, especially considering the black-box nature of many models. The confusion matrix is one of the core metric used to evaluate how well a classification model performs, on which many other metrics can be derived and computed: 
\vspace{-0.1cm}

\begin{itemize} 
\item True positive(TP): the number of positive samples predicted by the model to be positive class.
\item False positive(FP):the number of negative samples predicted by the model to be positive class.
\item True negative(TN): the number of negative samples predicted by the model to be negative class.
\item False negative(FN): the number of positive samples predicted by the model to be negative class.
\end{itemize}

FL requires continuous iteration, so it is necessary to demonstrate the classification performance of each iteration, including overall performance and classification performance in each class. For personalized solutions, it is also necessary to show the classification performance and variance of each participant.

We first introduce the metrics for evaluating the classification performance of single-class sample.

Precision is computed as the portion of true positives among all predicted positives. It reflects how accurate of a model in making positive predictions for one class. It is formulated as:
\vspace{-0.2cm}
\begin{equation}
Precision=\frac{TP}{TP + FP}
\end{equation}

Recall is computed as the portion of true positives among all ground-truth positves. It indicates how many positive samples can be correctly found by the model. It is formulated as follows:
\vspace{-0.2cm}
\begin{equation}
Recall=\frac{TP}{TP + FN}
\end{equation}

Sensitivity, also known as true positive rate (TPR), measures the classification accuracy of the samples belonging to positive class:
\vspace{-0.3cm}
\begin{equation}
Sensitivity=\frac{TP}{TP + FN}
\end{equation}

Specificity, also known as true negative rate (TNR), measures the classification accuracy of the samples belonging to negative class:
\vspace{-0.3cm}
\begin{equation}
Specificity=\frac{TN}{TN + FP}
\end{equation}

The false positive rate (FPR) describes the occurrence of false positives:
\vspace{-0.2cm}
\begin{equation}
FPR=\frac{FP}{FP + TN}
\end{equation}

The false negative rate(FNR) describes the occurrence of false negatives:
\vspace{-0.2cm}
\begin{equation}
FNR=\frac{FN}{FN + TP}
\end{equation}

Next, we introduce the evaluation metrics for evaluating the overall classification performance of all samples.

The F$_1$ score is the harmonic mean of precision and recall. The average F$_1$ score is calculated as:
\vspace{-0.1cm}
\begin{equation}
F_1=\frac{2*Precision*Recall}{Precision + Recall}
\end{equation}

The accuracy reflects the classification accuracy of all samples
\vspace{-0.2cm}
\begin{equation}
Accuracy=\frac{TP+TN}{TP + TN+FP+FN}
\end{equation}

Sometimes, accuracy is also termed as the top-1 accuracy, since the model's predicted label is picked as the class with the highest (top-1) probability from the model's final softmax layer. 

The G-means globally measures the classification ability of positive and negative samples
\vspace{-0.2cm}
\begin{equation}
G-means=\sqrt{Specificity \ast  Sensitivity}
\end{equation}

Matthews correlation coefficient (MCC) comprehensively evaluates the classification ability.
\vspace{-0.1cm}
\begin{equation}
MCC=\frac{TP \times TN-FP \times FN}{\sqrt{(TP+FP)(TP+FN)(TN+FP)(TN+FN)}}
\end{equation}

The receiver operating characteristics (ROC) curve maps the TPR to the FPR. The value of Area Under the ROC Curve (AUC) is the size of the area under the ROC curve. The larger the area, the more accurate the model classification. The AUC value of a general classifier is between 0.5 and 1: 0.5 means that there is no difference between the model's discriminating ability and random guessing; and 1 means the model is perfectly accurate. 
\vspace{-0.2cm}

\subsection{Classification Performance Evaluation of Federated Learning Model with Class Imbalance}
Evaluation of FL models with class imbalance is usually based on the metrics that are discussed above. However, it is crucial to evaluate the performance of FL model from different aspects to obtain a full picture of the model capability.  

\textbf{Training and testing performance}. The development of a ML model is usually divided into the training  (+validation) and testing stage. So does the FL model. The model performances of these two stages need to be evaluated in order to understand the FL model. Although a bit variance in model performance between training and testing stage is normal, similar performance in the two stages usually indicates that a well-trained model with good generalizability to unseen data. High performance during training with low performance on testing data may reflect the risk of over-fitting.  

\textbf{Global and client performance}. The global model obtained by aggregating all client models in FL may have different performance with each of the client model. Thus, the performance of the model on the server side (performance on the overall testing set) and the performance on each client side (performance on the client's local training set or testing set) need to be evaluated. In particular, in the personalized FL framework, multiple local models are produced, and hence, the performance of each user's local model needs to be evaluated.

\textbf{Majority class and minority class performance}. For the class imbalance problem, it is necessary to show the classification performance for each class, especially those minority classes.

\textbf{Iteration performance}. The training of an FL model requires many iterations. Thus, it is useful to track the model performance after each iteration and the number of training rounds required for the global model to converge.
\vspace{-0.2cm}

\section{Challenges And Future Directions}
\label{sec:challenges}
Although extensive research has been carried out to resolve the issue of class imbalance in FL, there still exist various open challenges that need to be addressed in the future. In this section, we discuss these open challenges and point out some possible future directions in this domain.

\textbf{Privacy protection.}
Although the client's data is not shared with the central server in FL, there are still potential risks for a malicious server to reconstruct a client's private data with or without additional assistance through the client's uploaded model. Existing solutions for safeguard user privacy in FL includes differential privacy \cite{huang2020dp}, secure multi-party computation \cite{du2004privacy} \cite{mohassel2017secureml}, and homomorphic encryption (HE) \cite{yuan2013privacy}. HE permits calculations to be conducted directly on ciphertexts and is widely adopted in FL frameworks to protect client's privacy. HE can be directly linked into existing FL solutions \cite{hardy2017private} \cite{liu2020secure} \cite{zhang2020batchcrypt}. For example, Paillier \cite{paillier1999public} is an established cryptosystem that has been implemented in FL systems, most notably FATE. The downside of HE is the increased transmission and encryption costs when the number of model parameters is huge \cite{zhang2020batchcrypt} \cite{zhang2021dubhe}.

With these, recent studies have shown that the protection of clients' privacy in FL can still be compromised \cite{geiping2020inverting}. One of the major causes of such compromise is through the model parameters (e.g., gradients), which may leak sensitive information to malicious adversaries and cause deep privacy leakage \cite{bhowmick2018protection} \cite{zhu2019deep}. As reported in \cite{aono2017privacy}, a small portion of the original gradients could reveal privacy about local training datasets. To solve the class imbalance problem, many existing methods discussed in this paper need to access the gradients of clients' local model, while overlook the risks of exposure of users' privacy to a certain extent. Therefore, it will be interesting if FL can further leverage homomorphic encryption or differential privacy techniques to protect user privacy via ``encrypting'' gradient and solve the class imbalance problem at the same time \cite{hahn2019privacy} \cite{yin2021comprehensive} \cite{mothukuri2021survey} \cite{aono2017privacy}. 

\textbf{Dynamic.}
Most FL methods are often modeled as a static application scenario where the data classes of the entire FL framework are fixed and predetermined \cite{dong2022federated}. However, real-world applications are usually dynamic with new data being generated everyday. Existing FL approaches often require storing all training data of old classes on the client side, but the high storage and computation costs may make FL unrealistic when new classes emerge dynamically \cite{nguyen2018crowdsourced} \cite{wang2021non} \cite{yang2021flop}. Moreover, if these approaches are required to learn new classes continuously with very limited storage memory \cite{shoham2019overcoming} \cite{yang2021flop}, they may suffer from significant performance degradation (i.e., catastrophic forgetting \cite{kirkpatrick2017overcoming} \cite{rebuffi2017icarl} \cite{shin2017continual}) in old classes. Finally, in real-world scenarios, new clients that collect the data of new classes in a streaming manner may want to participate in the FL training, which could further exacerbate the catastrophic forgetting on old classes in the global model training \cite{dong2022federated}. Therefore, it will be interesting for the community to come up with solutions that deal with the dynamic real-world scenarios.

\textbf{Data shift.}
Concept shift refers to the scenario where similar data samples are assigned with dissimilar labels by different clients, due to the clients' personal preferences. Such scenarios are quite common in real-world applications, especially the training data from all clients are horizontally overlapping, for example, image recognition, smart keyboards, and recommender systems, just to name a few. To give an example, when FL is used to train a recommender system using data contributed by various groups of people, concept shift will occur if the different participating groups of people have different cultural backgrounds with distinct preferences. Crowd-sourced data \cite{bassily2014private} is a more complex situation where data labels may be noisy, posing significant challenges to information collection in many traditional centralized ML tasks, let alone in FL tasks. For instance, some clients only have unlabeled data and need to have the data being labeled by several different people. Therefore, it is typical for some labels to be inaccurate, noisy, or even missing. In this situation, FL may fail to give a single global model that can fit data distributions of all clients \cite{fallah2020personalized} \cite{ghosh2020efficient} \cite{kairouz2021advances} \cite{mansour2020three} \cite{sattler2020clustered} \cite{smith2017federated} \cite{fu2021cic}.

\textbf{Relation between local models and global model.}	
In FL, to achieve high accuracy for the global model, it is important to keep the diversity among all clients' datasets. It has been found that distributing a lot of data with little class diversity may yield lower accuracy of global model while higher global performance can be achieved by distributing small numbers of data with diverse classes amongst some clients \cite{sittijuk2021performance}. This is because each client can collaboratively fulfill their data samples in each FL training process. Thus, adjusting the local imbalance on the client side may hurt the diversity amongst clients, and deteriorate the performance of global model. However, if there is a significant mismatch between the local imbalance and the global imbalance, the FL model's effectiveness will degrade. This poses a challenge for the global model to maintain a good classification performance for minority classes while keeping good classification performance for each client. It is also important to explore the relationship among the local imbalance, the global imbalance, and the mismatch imbalance, and explore the impact of three types of class imbalance on the classification performance of test dataset and clients' local datasets.

Given the above challenges in the class imbalanced FL, we suggest future directions as follows:
\begin{itemize}
\item It is necessary to further protect the client's privacy while solving the class imbalance problem. Most of the current methods require users to upload plaintext parameters such as gradients, which will still leak client privacy.

\item When dealing with dynamic class imbalanced data, how to ensure the classification performance of the old classes is a pressing issue that needs to be resolved.

\item For some FL application, the concept shift phenomenon among clients adds another dimension to the problem of class imbalance: the labels are not only skewed, but are also inconsistent and noisy amongst clients. This calls for FL methods that can accommodate the different preferences of clients, while addressing the class imbalance issue.

\item Local datasets owned by different clients may contain conflicted patterns. Therefore, obtaining a high performing local model for one client may not lead to a high performing global model sometimes. Hence, it will be interesting to propose methods that strike a balance between the global model, and the local model for each client. 
\end{itemize}
\vspace{-0.25cm}

\section{Conclusion}
\label{sec:conclusion}
This paper provides a systematic overview of FL with class imbalance. We introduce various types of class imbalance in FL systems and provide a comprehensive summary of existing techniques for handling class imbalance data. We start with a detailed categorization of class imbalance data in FL, after which we review the approaches proposed for class distribution estimation. Then, we provide a thorough review of existing works on handling class imbalance data in FL. We further present the metrics that are used to evaluate the reviewed approaches. Finally, we discuss the remaining challenges in this domain and suggest a few research directions to address these open questions.
\vspace{-0.25cm}

\ifCLASSOPTIONcompsoc
  \section*{Acknowledgments}  
  This work is supported by the National Key Research and Development Program of China (Grant No. 2021YFF0307103) and the National Natural Science Foundation of China under Grant 61872071 and Basic Scientific Research Business Expenses under Grant N2116010.
\else
  \section*{Acknowledgment}
\fi


\ifCLASSOPTIONcaptionsoff
  \newpage
\fi



\vspace{-0.2cm}
\bibliographystyle{IEEEtran}
\bibliography{mybibfile}
%
%
%

%

\begin{IEEEbiography}[{\includegraphics[width=1in,height=1.25in,clip,keepaspectratio]{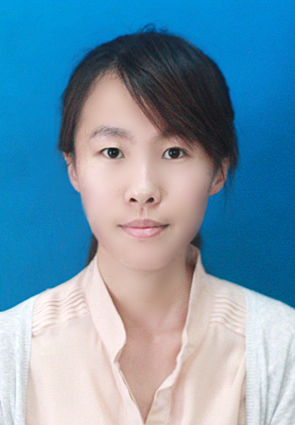}}]{Zhang Jing}
	received the B.S. and master’s degrees from Northeastern Forestry University and Dalian University of Technology, in 2015 and 2018, respectively. She is currently pursuing the Ph.D. degree in the School of Computer Science and Engineering at Northeastern University.
	
	Her current research interests include privacy preservation and machine learning.
\end{IEEEbiography}

\begin{IEEEbiography}[{\includegraphics[width=1in,height=1.25in,clip,keepaspectratio]{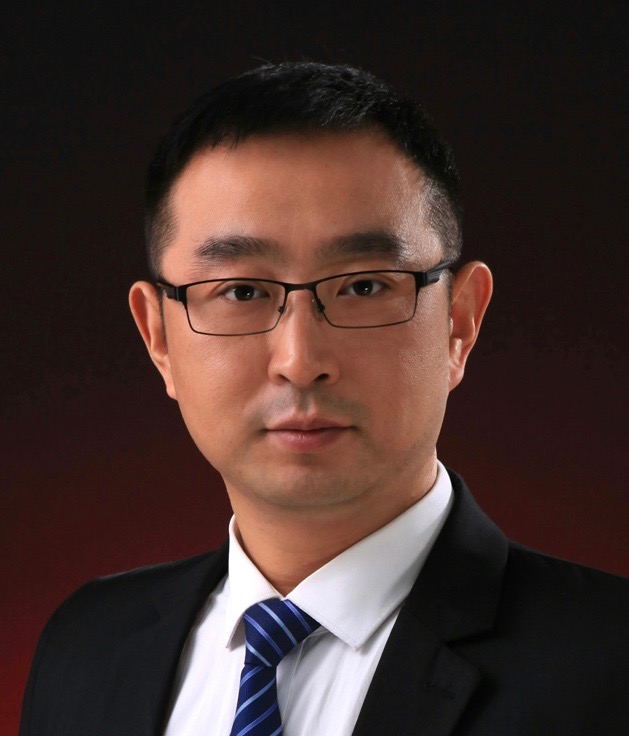}}]{Li Chuanwen}
	received the Ph.D. degree from Northeastern University, Shenyang, China, in 2011.
	
	He is currently an associate professor with Northeastern University. His current research interests include spatio-temporal data and location services, data management, distributed computing, cloud computing, and big data.
\end{IEEEbiography}

\begin{IEEEbiography}[{\includegraphics[width=1in,height=1.25in,clip,keepaspectratio]{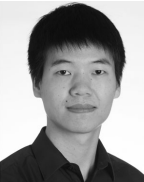}}]{Jianzhong Qi}
	 received the PhD degree from the University of Melbourne, Melbourne, Australia, in 2014. He is currently a senior lecturer with the School of Computing and Information Systems, University of Melbourne. His research interests include machine learning and data management and analytics, with a focus on spatial, temporal, and textual data.
\end{IEEEbiography}

\begin{IEEEbiography}[{\includegraphics[width=1in,height=1.25in,clip,keepaspectratio]{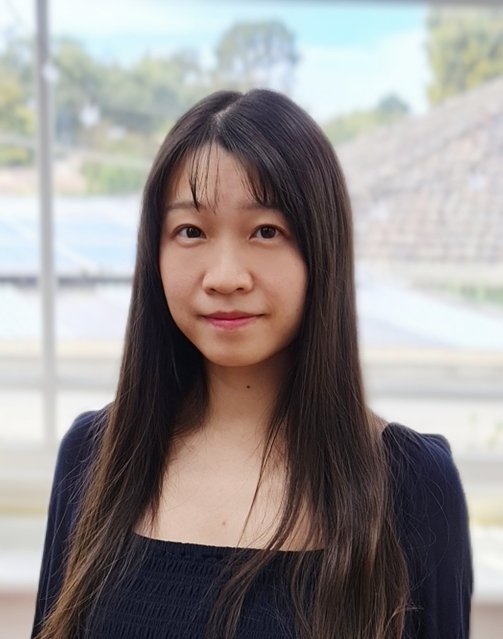}}]{Jiayuan He}
is a lecturer and early-career development fellow at RMIT University (School of Computing Technologies), Melbourne, Australia. She is also a member of the Research Centre for Information Discovery and Data Analytics. Before joining RMIT, she obtained her PhD from School of Computing and Information Systems at The University of Melbourne (UniMelb), followed by a postdoctoral research fellowship at UniMelb. Her research interests mainly focus on data mining and natural language processing.
\end{IEEEbiography}





\end{document}